%% file: colm2026_conference.tex
\documentclass{article} 
\usepackage[final]{colm2026_conference}

\usepackage{microtype}
\usepackage{hyperref}
\usepackage{url}
\usepackage{booktabs}
\usepackage{array}
\usepackage{graphicx}
\usepackage{multirow}
\usepackage{amsmath}
\usepackage{amssymb}
\usepackage{makecell}

\usepackage{lineno}

\definecolor{darkblue}{rgb}{0, 0, 0.5}
\hypersetup{colorlinks=true, citecolor=darkblue, linkcolor=darkblue, urlcolor=darkblue}
\newcommand{\para}[1]{

\noindent\textbf{#1}}

\title{Prescriptive Scaling Laws for Data Constrained Training}

\author{Justin Lovelace\thanks{Corresponding author: \texttt{jl3353@cornell.edu}}, Christian Belardi, Srivatsa Kundurthy, \\ \textbf{Shriya Sudhakar, Kilian Q. Weinberger} \\
Department of Computer Science \\
Cornell University \\
Ithaca, NY
}

\begin{document}

\ifcolmsubmission
\linenumbers
\fi

\maketitle

\input{sections/abstract}

\input{sections/introduction}

\input{sections/background}

\input{sections/experimental_setup}

\input{sections/method}

\input{sections/validation}

\input{sections/results}

\input{sections/discussion}

\input{sections/related_work}

\input{sections/conclusion}



\section*{Acknowledgments}
\input{sections/acknowledgment}


\bibliography{colm2026_conference}
\bibliographystyle{colm2026_conference}

\appendix
\input{sections/appendix_muennighoff}
\input{sections/appendix_training}
\input{sections/appendix_grid}
\input{sections/appendix_fits}
\input{sections/appendix_fwedu}

\input{sections/appendix_wd_sweep}
\input{sections/appendix_downstream}
\input{sections/appendix_llm}

\end{document}

%% file: sections/abstract.tex
\begin{abstract}
Training compute is increasingly outpacing the availability of high-quality data. This shifts the central challenge from optimal compute allocation to extracting maximum value from limited data. The widely adopted Chinchilla scaling law assumes every training token is unique. This limits its ability to guide pretraining decisions in data-constrained regimes. We model the excess loss under repetition with a simple additive overfitting penalty and find that it accurately describes model behavior. Our scaling law yields qualitatively new compute-optimal allocation advice. Beyond a point, further repetition is counterproductive and compute is better spent on model capacity. We show that following our law's recommended configuration improves performance in data-constrained regimes. Finally, because our one-parameter form isolates overfitting in a single coefficient, it enables direct comparison across training configurations. As a case study, we show that strong weight decay ($\lambda=1.0$) reduces this coefficient by approximately 70\%, providing a scaling-law explanation for recent findings that optimal weight decay in data-constrained regimes is an order of magnitude larger than standard practice. Our code is available at \url{https://github.com/justinlovelace/data-constrained-scaling}.

\end{abstract}

%% file: sections/introduction.tex
\section{Introduction}

Training compute is scaling faster than the supply of high-quality data. While raw text is abundant, the trend across state-of-the-art pipelines---aggressive quality filtering \citep{penedo2024fineweb}, upsampling of curated subsets \citep{olmo2025olmo}, and mid-training on domain-specific corpora \citep{allal2025smollm, olmo2025olmo}---reflects a new reality. Data, not compute, is the bottleneck. In specialized domains such as mathematics, code, and low-resource languages the constraint is even stronger. Domain-specific datasets  are often orders of magnitude smaller than the compute budget can absorb \citep{lewkowycz2022solving}. This shifts the central question from the Chinchilla framing of how to allocate compute optimally in the infinite-data regime \citep{hoffmann2022training} to \emph{how to extract the most from a fixed pool of data}, treating compute as effectively unbounded.

The simplest response, repeating data across multiple epochs, is already widespread. Yet the Chinchilla scaling law assumes every token is unique, and existing extensions \citep{muennighoff2023scaling}  for data repetition have a critical limitation. They can model diminishing returns but cannot represent the regime where loss \emph{increases} from overfitting. They also do not capture the interaction with overfitting and model size. In practice, larger models overfit faster on repeated data. Without explicitly modeling overfitting, we cannot accurately describe language modeling behavior under repetition.

We propose a simple additive overfitting penalty that encodes a simple intuition: overfitting is worse with limited data and larger models. We therefore model repeated tokens as useful while simultaneously incurring a separate, additive overfitting penalty that grows with repetitions.

We train hundreds of models spanning 15M--1B parameters, 50M--6B unique tokens, four weight decay strengths, and up to 16 epochs. Our central contribution is an additive overfitting penalty for data-constrained scaling laws that augments the Chinchilla law with a simple repetition term. A complexity ladder of 1-, 2-, and 4-parameter forms traces a Pareto frontier of fit quality versus complexity, with even the one-parameter form substantially outperforming prior data-constrained scaling laws. The fitted law yields qualitatively new compute-optimal allocation advice: beyond a data-dependent threshold, further repetition is counterproductive and compute is better spent on model capacity. We validate this prescription by training the configuration each law recommends and show that ours achieves the strongest performance in both perplexity and downstream evaluation. Finally, because our law isolates overfitting in a single coefficient, it enables direct comparison across training configurations. As a case study, we show that strong weight decay ($\lambda = 1.0$) reduces overfitting by approximately 70\%, and our law predicts the compute budget at which strong regularization overtakes the standard setting, demonstrating that it effectively guides training decisions in data-constrained regimes.

%% file: sections/background.tex
\section{Background and Related Work}

\para{Neural scaling laws.}
Training a large language model requires two key decisions: how big to make the model and how much data to train it on. Together these determine both how much \textit{compute} is needed and the \textit{performance} of the final model. Typically, performance is measured by the model's loss on held-out text. A natural question is how loss changes as we scale up the model or the dataset. Empirically, the answer is remarkably clean: loss falls predictably as either quantity grows, following a smooth mathematical trend that holds across many orders of magnitude. These trends are called \emph{neural scaling laws}, and their predictability is what makes them useful. Instead of training many large models to find the best configuration, which would be prohibitively expensive, a researcher can fit a scaling law to many small, cheap training runs. A scaling law can then be used to forecast the loss of a much larger run before committing the compute to it.

\citet{kaplan2020scaling} established these trends for large language models, and \citet{hoffmann2022training} refined them into the Chinchilla scaling law, which expresses the final loss of a trained model as a sum of three terms:
\begin{equation}
    L(N, D) = E + \frac{A}{N^\alpha} + \frac{B}{D^\beta},
    \label{eq:chinchilla}
\end{equation}
where $N$ is the number of model parameters and $D$ is the number of training tokens. Each of the three terms has a natural interpretation. The first term, $E$, is a floor that the loss cannot drop below no matter how much we scale compute. Natural language has inherent unpredictability (the next word is rarely fully determined by what came before), and $E$ captures that uncertainty. The second term, $A / N^\alpha$, represents the cost of having a model that is too small. A model with limited parameters cannot represent all the patterns present in language. As $N$ grows, this term shrinks toward zero, and the rate at which it shrinks is controlled by the exponent $\alpha$. The third term, $B / D^\beta$, represents the cost of having trained on too little data. Even a very large model has mediocre performance if it has not seen enough data. As $D$ grows, this term also shrinks toward zero, with its rate governed by the exponent $\beta$. Together, the three terms say that loss is the sum of what is fundamentally unlearnable, what the model was too small to learn, and what the data was too limited to teach.

A practical consequence of this form is that a compute budget determines how to trade off model size against dataset size. Training compute scales as $C \approx 6ND$ (see \citep{kaplan2020scaling}), so for a fixed $C$, the scaling law picks out a specific $(N, D)$ pair that minimizes predicted loss. The exact $N:D$ ratio depends on the fitted exponents and varies across scaling studies with different datasets and training setups~\citep{li2025misfitting}. 

Crucially, the Chinchilla law assumes that every training token is unique---the model sees each piece of text exactly once. In practice, high-quality text is often scarce enough that training runs pass over the same data multiple times, violating this assumption. Scaling laws of this form have been widely adopted for guiding training
decisions across domains and applications~\citep{ludziejewski2024scaling,
gulrajani2023likelihood, cherti2023reproducible, goyal2024scaling,
kumar2025scaling, aghajanyan2023scaling, sardana2024beyond, gadre2025language}.

\para{Scaling under data repetition.}
The Chinchilla law assumes each training token is seen exactly once, but in
practice models often train for multiple epochs over the same data. While
repeating data improves performance, each additional pass yields progressively
diminishing returns. At some point, additional epochs contribute almost
nothing, or can even hurt performance due to overfitting. Any scaling law that
accounts for repetition must capture this behavior. Recent work has begun to
characterize this regime: \citet{ni2025diffusion} show that repetition
tolerance depends on the training objective, and concurrent work examines overfitting when a small domain corpus is upsampled within a larger
pretraining mixture~\citep{baek2026finetuner}.

\citet{muennighoff2023scaling} formalized the diminishing-returns intuition by
replacing the raw token count $D$ in the Chinchilla law with an \emph{effective data} quantity $\widehat{D}$. The main idea is that the contribution of repeated tokens decays exponentially. The first repeat is worth almost as much as fresh data, the second repeat somewhat less, and so on, until further repeats add essentially nothing. Let $U_D$ denote the number of unique tokens in the dataset and let $R_D$ denote the number of \emph{additional} epochs beyond the first ($R_D = 0$ means the model sees each token once, $R_D = 1$ means twice, and so on). They define:
\begin{equation}
    \widehat{D}(U_D, R_D) = U_D \cdot \left(1 + R_D^* \cdot \left(1 - e^{-R_D / R_D^*}\right)\right),
    \label{eq:muennighoff_deff}
\end{equation}
where $R_D^*$ is a fit constant that controls how quickly repeated data loses its marginal value. The behavior of this expression matches the decay intuition directly. When $R_D$ is small, the exponential term is approximately linear, so $\widehat{D}(U_D, R_D) \approx U_D \cdot (1 + R_D)$---each repeated epoch contributes nearly as much as a fresh one. As $R_D$ grows, the exponential saturates toward one, so $\widehat{D}(U_D, R_D)$ approaches an upper limit of $U_D \cdot (1 + R_D^*)$. No matter how many more epochs are added, effective data cannot exceed this ceiling. Substituting $\widehat{D}$ into the Chinchilla form gives:
\begin{equation}
    L(N, U_D, R_D) = E + \frac{A}{N^\alpha} + \frac{B}{\widehat{D}(U_D, R_D)^\beta}.
    \label{eq:muennighoff_loss}
\end{equation}
However, this formulation treats repetition as purely a data-side phenomenon with no dependence on model size. Empirically, this is incomplete: larger models overfit more quickly on repeated data than smaller ones do, so the cost of repetition should depend on $N$ as well as $D$. To capture this, \citet{muennighoff2023scaling} apply the same saturating form to model parameters. The intuition is that a model can be too large for its dataset. If the parameter count far exceeds what the available unique tokens can support, the extra capacity yields diminishing returns. They measure excess capacity relative to $N_{\text{opt}}$, the Chinchilla compute-optimal model size for $U_D$ unique tokens. If the actual model size $N$ exceeds $N_{\text{opt}}$, the ratio $R_N = (N / U_N) - 1$ measures the degree of overparameterization; otherwise $R_N = 0$. Writing $U_N = \min\{N_{\text{opt}}, N\}$, the effective parameter count is:
\begin{equation}
    \widehat{N}(U_N, R_N) = U_N + U_N \cdot R_N^* \cdot \left(1 - e^{-R_N / R_N^*}\right),
    \label{eq:muennighoff_neff}
\end{equation}
where $R_N^*$ plays the same role as $R_D^*$ but for excess parameters rather than repeated tokens. Substituting both $\widehat{D}$ and $\widehat{N}$ into the Chinchilla form yields a scaling law with two additional fit constants ($R_D^*$ and $R_N^*$):
\begin{equation}
    L(U_N, R_N, U_D, R_D) = E + \frac{A}{\widehat{N}(U_N, R_N)^\alpha} + \frac{B}{\widehat{D}(U_D, R_D)^\beta}.
    \label{eq:muennighoff_full}
\end{equation}
This introduces an interaction between model size and repetition: overparameterized models see their effective capacity $\widehat{N}(U_N, R_N)$ saturate, which raises the predicted loss under heavy repetition. 

The mechanism is indirect, modeling overparameterization as a diminishing return on effective model size rather than as an explicit overfitting cost, and it is not clear why excess parameters should follow the same exponential saturation form as repeated data. In~\autoref{sec:method}, we propose an alternative formulation that explicitly separates the contribution of repeated tokens from the overfitting penalty they incur, and show that this decomposition reveals how regularization strength modulates a model's tolerance to repetition.

%% file: sections/experimental_setup.tex
\section{Experimental setup}
\label{sec:setup}

We pretrain decoder-only language models using the Llama~2 architecture and tokenizer~\citep{touvron2023llama} across a grid of model sizes, unique data budgets, and repetition counts.
All models are trained on the FineWeb dataset~\citep{penedo2024fineweb}, a large-scale filtered web corpus.
For our scaling study, we sweep over model sizes $N$ ranging from 15M to 1B parameters, unique data budgets $U_D$ from 50M to 6B tokens, and repetition counts $R_D \in \{0, 1, 3, 7, 11, 15\}$; the full experimental grid is detailed in~\autoref{app:experimental_grid}.
All configurations are trained at two weight decay strengths: the standard setting $\lambda = 0.1$ and a strong setting $\lambda = 1.0$ (extended sweep in \autoref{app:wd_sweep}). All other hyperparameters---learning rate, warmup schedule, batch size---are held constant across weight decay conditions, isolating the effect of regularization on repetition overfitting.
We report the final validation loss at the end of training.
Detailed model architectures, hyperparameter configurations, and training procedures are provided in~\autoref{app:training_details}. We evaluate our models with validation perplexity and also report downstream performance with the Open Language Model Evaluation Standard (OLMES) \citep{gu2025olmes}. We report the average bits-per-byte (BPB) across 19 downstream language understanding tasks from the recommended evaluation suite for small models \citep{heineman2025signal}.

%% file: sections/method.tex
\section{Scaling laws for repeated data}
\label{sec:method}

\begin{figure}[t]
    \centering
    \includegraphics[width=\textwidth]{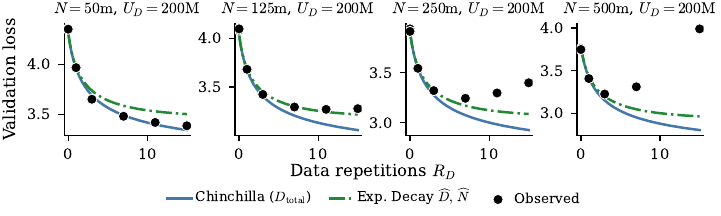}
    \caption{\textbf{Existing scaling laws fail to model overfitting.} Predicted vs.\ observed validation loss under the Chinchilla baseline (treating repeated tokens as unique) and the \citet{muennighoff2023scaling} $\widehat{D}(U_D, R_D), \widehat{N}(U_N, R_N)$ formulation across four model sizes at $U_D = 200$M. Both formulations fail to capture the loss increase at high repetition counts, systematically underpredicting loss as repetitions grow.}
    \label{fig:muennighoff_residuals}
    \vspace{-1em}
\end{figure}

\para{Limitations of the effective-data approach.} We begin by examining where the \citet{muennighoff2023scaling} formulation breaks down. Figure~\ref{fig:muennighoff_residuals} plots predicted vs.\ observed validation loss under the Chinchilla baseline and the $\widehat{D}(U_D, R_D), \widehat{N}(U_N, R_N)$ form across four model sizes. Two patterns emerge: first, the gap between predicted and observed loss grows with the number of repetitions, indicating that the effective-data forms systematically underpredict loss at high epoch counts. Second, the gap increases with model capacity relative to unique data, confirming an interaction between model size and repetition that the exponential saturation in $\widehat{D}(U_D, R_D)$ and $\widehat{N}(U_N, R_N)$ cannot capture.

\begin{figure}[t]
    \centering
    \includegraphics[width=\textwidth]{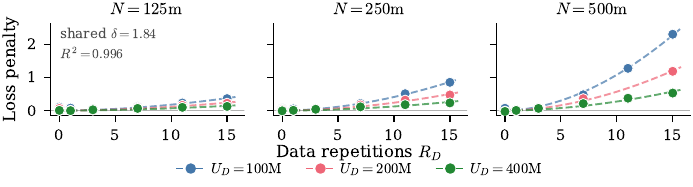}
    \caption{\textbf{The cost of repeating data grows superlinearly.} Residual between observed loss and the Chinchilla prediction (treating all repeated tokens as unique data) as a function of repetition count $R_D$, shown for three model sizes and three unique data budgets $U_D$. Dashed lines show power-law fits with a shared exponent $\delta$ across all configurations; the fitted $\delta > 1.0$ indicates superlinear repetition damage. Slopes are steeper for larger models and smaller unique data budgets.}
    \label{fig:residual_power}
    \vspace{-1em}
\end{figure}

\para{An additive overfitting penalty.} To identify the right functional form, we fit the Chinchilla law (Equation~\ref{eq:chinchilla}) to single-epoch runs, obtaining parameters $(E, A, B, \alpha, \beta)$, and use this fit to predict multi-epoch loss by treating repeated tokens as if they were fresh data $D = U_D \cdot (1 + R_D)$. The residual between the observed loss and this prediction isolates the additional cost attributable to repetition.
When we plot this residual for a fixed model size $N$ and unique data budget $U_D$, varying only the number of repetitions $R_D$, a power-law relationship emerges (Figure~\ref{fig:residual_power}). A shared power-law fit $P_i \cdot R_D^\delta$ across all (model, budget) configurations, with $\delta$ tied and $P_i$ free per cell, finds the repetition damage is \emph{superlinear}. Each additional epoch of repetition inflicts more damage than the last.

\para{A complexity ladder of penalty forms.} Examining how the per-cell coefficient $P_i$ varies across configurations reveals its structure: larger models and smaller unique data budgets incur steeper penalties. This motivates a family of additive penalty forms of increasing complexity, which we present as a Pareto frontier trading off number of free parameters against fit quality.
The simplest form uses a single free parameter $P$ and the dimensionless ratio $N / U_D$:
\begin{equation}
    L(N, U_D, R_D) = E + \frac{A}{N^\alpha} + \frac{B}{\left(U_D \cdot (1 + R_D)\right)^\beta} + P \cdot R_D \cdot \frac{N}{U_D}.
    \label{eq:one_param}
\end{equation}
This linear-in-$R_D$ form already substantially outperforms the \citet{muennighoff2023scaling} formulations (Section~\ref{sec:validation}).
Adding a second free parameter, the exponent $\kappa$ on the capacity ratio, allows the penalty to scale nonlinearly with model size relative to data:
\begin{equation}
    L(N, U_D, R_D) = E + \frac{A}{N^\alpha} + \frac{B}{\left(U_D \cdot (1 + R_D)\right)^\beta} + P \cdot R_D \cdot \left(\frac{N}{U_D}\right)^\kappa.
    \label{eq:two_param}
\end{equation}
Across configurations, $\kappa > 1$, indicating that the overfitting penalty grows superlinearly with the ratio of model capacity to unique data.
The full four-parameter form adds a superlinear exponent $\delta$ on the repetition count and decouples the data-budget exponent $\gamma$ from the model-size exponent $\kappa$:
\begin{equation}
    L(N, U_D, R_D) = E + \frac{A}{N^\alpha} + \frac{B}{\left(U_D \cdot (1 + R_D)\right)^\beta} + P \cdot R_D^\delta \cdot \left(\frac{N}{U_D^\gamma}\right)^\kappa.
    \label{eq:four_param}
\end{equation}

At $R_D = 0$ (single epoch), all three forms reduce exactly to the Chinchilla law. The key conceptual difference from the effective-data approach is that repeated tokens play a dual role: they continue to reduce the data-sufficiency term (they are not wasted) while simultaneously incurring a growing overfitting cost. 

%% file: sections/validation.tex
\section{Scaling law validation}
\label{sec:validation}

\begin{figure}[t]
    \centering
    \includegraphics[width=\textwidth]{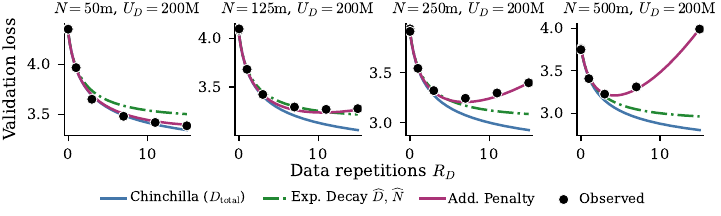}
    \caption{\textbf{Our additive overfitting law adapts to model size.} Predicted vs.\ observed validation loss under different scaling laws. }
    \label{fig:our_law_fit}
    \vspace{-1em}
\end{figure}

\begin{figure}[t]
    \centering

    \includegraphics[width=\textwidth]{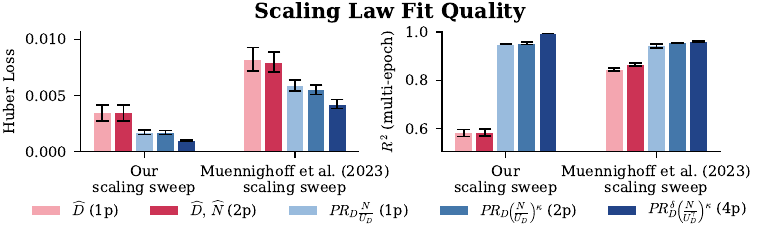}
    \caption{\textbf{Scaling law fit quality.} Huber loss (left, lower is better) and $R^2$ (right, higher is better) for each scaling law formulation, evaluated on both our scaling sweep and the \citet{muennighoff2023scaling} data. }
    \label{fig:quantitative_r2}
    \vspace{-1em}
\end{figure}

We validate the additive penalty laws (\autoref{eq:one_param}--\autoref{eq:four_param}) against the Chinchilla baseline and the \citet{muennighoff2023scaling} effective-data formulations on two independent scaling sweeps. The first is our own CLM sweep described in \autoref{sec:setup}. The second is the public scaling sweep from \citet{muennighoff2023scaling}, including all runs up to 64 epochs—a more lenient filter than the outlier-removal criteria applied in their original analysis.

\autoref{fig:our_law_fit} compares predicted and observed loss across model sizes. Our proposed law accurately tracks the observed loss across all configurations, capturing the degradation under heavy repetition that the effective-data formulations miss. ~\autoref{fig:quantitative_r2} confirms the quantitative improvement: even our one-parameter additive penalty (\autoref{eq:one_param}) substantially outperforms both the $\widehat{D}(U_D, R_D)$ and $\widehat{D}(U_D, R_D), \widehat{N}(U_N, R_N)$ forms, and the four-parameter form (\autoref{eq:four_param}) achieves near-perfect fit on our data. The improvement extends to the heldout \citet{muennighoff2023scaling} data, where model sizes and repetition ranges span a wider range.

\begin{figure}[t]
    \centering
    \includegraphics[width=\textwidth]{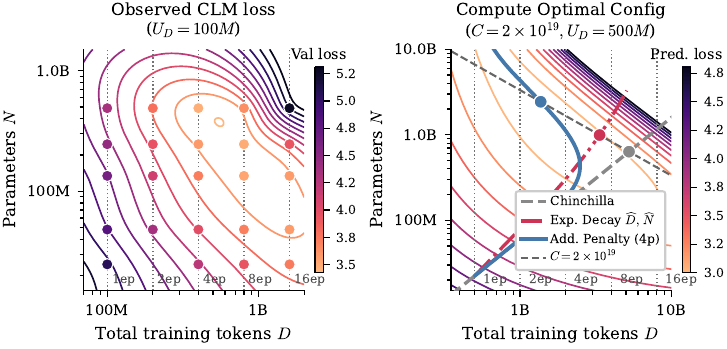}%
    \caption{\textbf{Compute-optimal allocation frontiers.} \textbf{(Left)}~Observed validation loss across our experimental grid ($U_D = 100$M), with contour lines showing the interpolated loss landscape. \textbf{(Right)}~Compute-optimal allocation at $C = 2 \times 10^{19}$ FLOPs and $U_D = 500$M, comparing the Chinchilla law, \citet{muennighoff2023scaling} $\widehat{D}, \widehat{N}$, and our law (\autoref{eq:four_param}).}
    \label{fig:allocation_frontier}
    \vspace{-1em}
\end{figure}

\para{Compute-optimal allocation under repetition.} The superlinear repetition penalty yields qualitatively different compute-optimal allocation advice than prior scaling laws (\autoref{fig:allocation_frontier}). The Chinchilla law, which ignores overfitting, always recommends more repetition: the optimal total token count $D$ grows linearly with compute at fixed $U_D$. The \citet{muennighoff2023scaling} $\widehat{D}(U_D, R_D),\widehat{N}(U_N, R_N)$ form moderates this, prescribing diminishing returns from repetition but never recommending that repetition stop. Our four-parameter law, by contrast, predicts a compute budget beyond which additional repetition is counterproductive. The allocation frontier \emph{turns back}: at high compute, the law recommends scaling model size while \emph{reducing} the number of epochs\footnote{On its surface, this recommendation appears to contradict \citet{muennighoff2023scaling}, who find that data-constrained compute should be allocated toward smaller models trained for more epochs. We trace this disagreement to a methodological choice in their analysis; see \autoref{app:muennighoff_reanalysis}.}, reflecting the overfitting cost of continued repetition. This provides concrete guidance for practitioners: given a fixed data budget, there is a compute level beyond which training a larger model for fewer epochs outperforms training a smaller model for more.

\para{Prescriptive validation.} The $R^2$ comparisons above measure descriptive fit---how well each law explains configurations it was trained on. A more stringent test is \emph{prescriptive} accuracy: given a fixed unique-data and compute budget, does the law recommend the configuration that actually achieves the lowest loss? For each (token budget, compute budget) pair in \autoref{tab:held_out_validation}, we solve each law for the optimal model size and epoch count, train the recommended configuration, and evaluate. Our law consistently recommends larger models with fewer epochs and achieves the best perplexity and downstream performance across all settings.

\begin{table}[t]
    \centering
    \caption{\textbf{Prescriptive validation.} For each (token budget, compute budget) pair, we train the configuration recommended by each scaling law: Chinchilla \citep{hoffmann2022training}, Eff.\ Param.\ \citep{muennighoff2023scaling}, and Ours (\autoref{eq:four_param}). OLMES BPB is the average bits-per-byte across 19 downstream tasks in the OLMES evaluation harness \citep{gu2025olmes}.}
    \label{tab:held_out_validation}
    \begin{tabular}{llllrrr}
        \toprule
        $U_D$ & $C$ & Scaling law & Params & Epochs & Perplexity $\downarrow$ & OLMES BPB $\downarrow$ \\
        \midrule
        \multirow{3}{*}{250M}
        & \multirow{3}{*}{$5 \times 10^{18}$}
        & Chinchilla & 280M & 12 & 25.31 & 1.52 \\
        & & Eff.\ Param.\ & 500M & 7 & 23.91 & 1.50 \\
        & & Ours & 700M & 5 & {22.90} & {1.45} \\
        \midrule
        \multirow{6}{*}{500M}
        & \multirow{3}{*}{$1 \times 10^{19}$}
        & Chinchilla & 390M & 8 & 18.95 & 1.35 \\
        & & Eff.\ Param.\ & 550M & 6 & 18.65 & 1.36 \\
        & & Ours & 700M & 5 & {18.48} & {1.30} \\
        \cmidrule(l){2-7}
        & \multirow{3}{*}{$2 \times 10^{19}$}
        & Chinchilla & 700M & 10 & 20.44 & 1.38 \\
        & & Eff.\ Param.\ & 950M & 7 & 19.34 & 1.40 \\
        & & Ours & 2.2B & 3 & {17.73} & {1.34} \\
        \bottomrule
    \end{tabular}
    \vspace{-1em}
\end{table}

\para{Generalization to external data.} To test the generalizability of our scaling law form, we apply it to the published data from \citet{muennighoff2023scaling}, which was completely held out during the development of our scaling law. It therefore represents a true test of generalization. They make different choices with respect to backbone architecture (GPT-2 versus Llama-2), tokenization, etc. for pre-training. We find that the additive penalty forms significantly  outperforms the \citet{muennighoff2023scaling} formulations on their own published data, confirming that the improvement generalizes across pre-training implementations.

\para{Generalization across datasets.}
We re-run the full sweep on the aggressively filtered FineWeb-Edu
corpus (full details and results in \autoref{app:fineweb_edu}). We re-fit the scaling laws and find that this sweep reproduces the fit-quality rankings from \autoref{fig:quantitative_r2} exactly. Both effective-data
baselines struggle to model multi-epoch loss
($R^2_{\text{multi}} \approx 0.63$) while every additive form dominates
($0.95$--$0.99$), and overfitting sets in at comparable repetition counts.

%% file: sections/results.tex
\section{Case study: weight decay improves robustness to data repetition}
\label{sec:results}

The overfitting coefficient directly quantifies a training configuration's robustness to data repetition. To demonstrate this, we analyze the effect of weight decay strength on repetition tolerance, training the same grid of model sizes and data budgets at two settings: standard $\lambda = 0.1$ and strong $\lambda = 1.0$ weight decay, with all other hyperparameters held constant.

\begin{figure}[t]
    \centering
    \includegraphics[width=\textwidth]{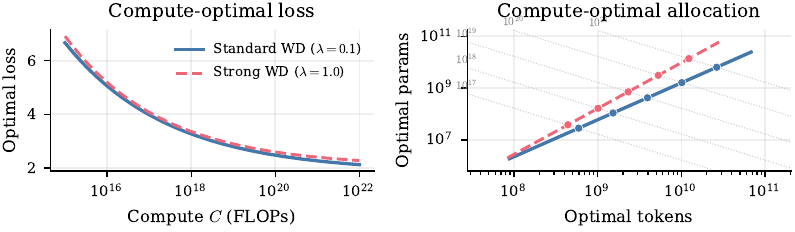}
    \caption{\textbf{Strong weight decay incurs a single-epoch loss premium.} \textbf{(Left)}~Compute-optimal frontier: strong weight decay ($\lambda=1.0$) achieves higher loss than the standard setting at every compute budget in the single-epoch regime. \textbf{(Right)}~Compute-optimal allocation: strong weight decay favors larger models relative to data.}
    \label{fig:single_epoch_wd}
    \vspace{-.5em}
\end{figure}

\para{Single-epoch scaling.} We fit separate Chinchilla parameters $(A, B, E, \alpha, \beta)$ per setting. In the single-epoch regime, strong weight decay incurs a loss premium at every compute budget (\autoref{fig:single_epoch_wd}), and its compute-optimal allocation favors larger models relative to data.

\begin{figure}[t]
    \centering
    \includegraphics[width=\textwidth]{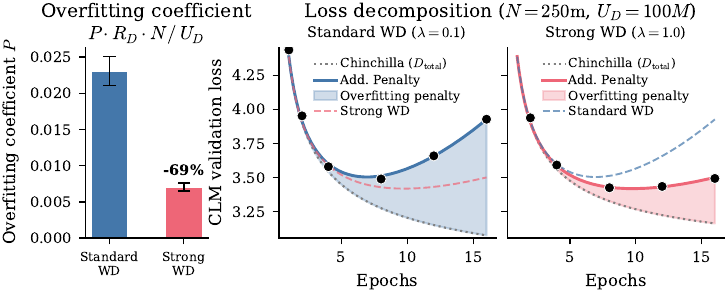}
    \caption{\textbf{Strong weight decay improves robustness to data repetition.} \textbf{(Left)}~Fitted overfitting coefficient $P$ for standard ($\lambda=0.1$) and strong ($\lambda=1.0$) weight decay; strong weight decay reduces $P$ by approximately 70\%. \textbf{(Center, Right)}~Loss decomposition for a 250M-parameter model trained on 100M unique tokens across weight decay values.}
    \label{fig:loss_decomposition}
    \vspace{-1em}
\end{figure}

\para{Scaling under data repetition.} We fit our additive penalty (\autoref{eq:four_param}) independently for both weight decay settings.
\autoref{fig:loss_decomposition} (left) shows the fitted $P$ values from the one-parameter form (\autoref{eq:one_param}): strong weight decay reduces $P$ by approximately 70\%, meaning it incurs far less overfitting per repetition. The loss decomposition for a representative configuration (center, right) shows the overfitting penalty growing superlinearly for both settings, but with significantly lower magnitude under strong weight decay.

\begin{figure}[t]
    \centering
    \includegraphics[width=\textwidth]{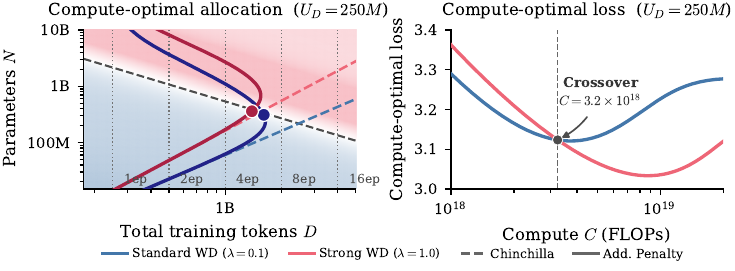}
    \caption{\textbf{Strong weight decay outperforms in high-compute regimes.} \textbf{(Left)}~Compute-optimal allocation frontiers for standard ($\lambda=0.1$) and strong ($\lambda=1.0$) weight decay at $U_D = 250$M. \textbf{(Right)}~Compute-optimal loss as a function of compute budget. Strong weight decay incurs higher loss at low compute but crosses over at $C \approx 3.2 \times 10^{18}$ FLOPs.}
    \label{fig:crossover_multi}
    \vspace{-1em}
\end{figure}

\para{Crossover under repetition.} Although there is a single-epoch loss premium, the significantly reduced overfitting cost creates a crossover point in performance (\autoref{fig:crossover_multi}). At $U_D = 250$M, standard weight decay achieves lower loss at modest compute, but strong weight decay overtakes it at $C \approx 3.2 \times 10^{18}$ FLOPs as the standard setting's steeper penalty erodes its single-epoch advantage. Our scaling law predicts this crossover point: the compute budget at which the lower penalty compensates for the single-epoch tax.

\begin{table}[t]
    \centering
    \caption{\textbf{Weight decay prescriptive validation.} For each (token budget, compute budget) pair, we train the configuration recommended by our scaling law (\autoref{eq:four_param}). Standard weight decay ($\lambda = 0.1$) achieves lower loss at small compute budgets, but strong weight decay ($\lambda = 1.0$) enables productive use of additional compute, ultimately achieving lower absolute loss. We denote the expected crossover point under our law with $C^{\times}$.}
    \label{tab:wd_validation}
    \begin{tabular}{llllrrr}
        \toprule
        $U_D$ & $C$ & Weight Decay & Params & Epochs & Perplexity $\downarrow$ & BPB  $\downarrow$\\
        \midrule
        \multirow{6}{*}{\makecell[l]{250M\\[0pt]{\small $C^{\times}\! \approx 3\!\times\!10^{18}$}}}
        & \multirow{2}{*}{\textcolor{gray}{$3 \times 10^{18}$}}
        & \textcolor{gray}{$\lambda = 0.1$} & \textcolor{gray}{350M} & \textcolor{gray}{6} & \textcolor{gray}{23.38} & \textcolor{gray}{1.46} \\
        & & \textcolor{gray}{$\lambda = 1.0$} & \textcolor{gray}{350M} & \textcolor{gray}{6} & \textcolor{gray}{22.93} & \textcolor{gray}{1.47} \\
        \cmidrule(l){2-7}
        & \multirow{2}{*}{$5 \times 10^{18}$}
        & $\lambda = 0.1$ & 700M & 5 & 22.91 & 1.47 \\
        & & $\lambda = 1.0$ & 550M & 6 & {21.66} & {1.41} \\
        \cmidrule(l){2-7}
        & \multirow{2}{*}{$1 \times 10^{19}$}
        & $\lambda = 0.1$ & 3B & 2 & 23.13 & 1.52 \\
        & & $\lambda = 1.0$ & 1B & 6 & {20.34} & {1.36} \\
        \midrule
        \multirow{4}{*}{\makecell[l]{500M\\[0pt]{\small $C^{\times}\!\approx 1\!\times\!10^{19}$}}}
        & \multirow{2}{*}{\textcolor{gray}{$1 \times 10^{19}$}}
        & \textcolor{gray}{$\lambda = 0.1$} & \textcolor{gray}{700M} & \textcolor{gray}{5} & \textcolor{gray}{18.52} & \textcolor{gray}{1.36} \\
        & & \textcolor{gray}{$\lambda = 1.0$} & \textcolor{gray}{830M} & \textcolor{gray}{4} & \textcolor{gray}{18.75} & \textcolor{gray}{1.32} \\
        \cmidrule(l){2-7}
        & \multirow{2}{*}{$3 \times 10^{19}$}
        & $\lambda = 0.1$ & 5B & 2 & 18.16 & 1.35 \\
        & & $\lambda = 1.0$ & 2.5B & 4 & {16.63} & {1.30} \\
        \bottomrule
    \end{tabular}
    \vspace{-1.5em}
\end{table}

\para{Prescriptive validation.} We validate on held-out configurations (\autoref{tab:wd_validation}). The results confirm the crossover dynamics predicted by our law. Near the predicted crossover budget $C^{\times}$, the two settings perform comparably. Beyond this point, the gap widens rapidly---at $C = 1 \times 10^{19}$ (${\sim}3\times$ past the crossover), strong weight decay reduces perplexity by 2.8 points. The same pattern holds at $U_D = 500$M: near the crossover the settings are comparable, but at $3 \times 10^{19}$ strong weight decay achieves 16.63 vs.\ 18.16 perplexity. Our law correctly predicts that strong weight decay should train for more epochs, and the recommended configurations achieve lower loss when the data constraint is binding. This provides prescriptive support for the empirical finding of \citet{kim2026pretraining} that optimal weight decay in data-constrained regimes can be an order of magnitude larger than standard practice.

%% file: sections/discussion.tex
\para{Practical implications.} The compute-optimal allocation frontier (\autoref{fig:allocation_frontier}) provides concrete guidance for data-constrained practitioners. Given a fixed data budget, there is a compute threshold beyond which repeating data further is counterproductive. Past this threshold, compute is better allocated to model capacity. Combined with the finding that strong weight decay reduces the overfitting coefficient $P$ by approximately 70\%, practitioners in data-constrained settings have two complementary levers: increasing regularization strength and choosing the right model size--epoch tradeoff.

\para{Beyond two settings.} In \autoref{app:wd_sweep} we extend the
sweep to $\lambda \in \{0.1, 0.5, 1.0, 3.0\}$. The overfitting coefficient
$P$ falls smoothly and monotonically with $\lambda$, but stronger regularization also raises the single-epoch
loss, so the optimal $\lambda$ is regime-dependent. The compute-optimal weight decay
value increases
with the compute budget and with tighter data constraints. Our law predicts the
crossover budget at which each setting overtakes the last.

\para{Limitations.} Our study spans model sizes up to 1B parameters and repetition counts up to 16 epochs; we cannot confirm that the fitted exponents hold at the frontier scales used in practice. Our scaling law form may not extrapolate to extreme repetition counts beyond those covered in our study. Notably, our law doesn't encapsulate phenomena such as double descent \citep{Nakkiran2020Deep}. We fit an independent scaling sweep per weight decay setting rather than modeling $\lambda$ as a continuous input, so recommending a $\lambda$ outside the four we test requires interpolating between fits. Extending our law to incorporate regularization strength directly is an interesting direction for future work.

%% file: sections/related_work.tex

%% file: sections/conclusion.tex
\section{Conclusion}

We have presented a simple data-constrained scaling law that models the cost of data repetition with a simple, additive overfitting penalty. A complexity ladder of one-, two-, and four-parameter forms provides a Pareto frontier of fit quality versus complexity, with even the simplest form substantially outperforming prior effective-data formulations.
The overfitting penalty provides a new axis for evaluating training configurations, directly quantifying robustness to data repetition. This property is of increasing importance as compute grows faster than data. As a case study, we show that strong weight decay ($\lambda=1.0$) reduces the overfitting penalty by approximately 70\%, providing prescriptive guidance that explains, and can replace, the expensive per-configuration hyperparameter sweeps required by prior empirical work~\citep{kim2026pretraining}.
Finally, our overfitting penalty yields qualitatively new compute-optimal allocation advice: beyond a threshold compute budget, further data repetition is counterproductive, and practitioners should scale model capacity instead. Together, these findings provide concrete guidance for the increasingly common setting where high-quality data, not compute, is the binding constraint.

%% file: sections/acknowledgment.tex
JL is supported by a Google PhD Fellowship.
This work is supported by AI-MI and NSF Award DMR-2433348, and by NSF Award IIS-2530143. We gratefully acknowledge use of the research computing resources of the Empire AI Consortium, Inc., with support from the State of New York, the Simons Foundation, and the Secunda Family Foundation.

%% file: sections/appendix_muennighoff.tex
\newpage
\section{Reanalysis of Muennighoff et al.\ compute-optimal allocation}
\label{app:muennighoff_reanalysis}

Our scaling law recommends allocating data-constrained compute toward larger models trained for fewer epochs, while \citet{muennighoff2023scaling} recommend the opposite: smaller models trained for more epochs. We trace this disagreement to the fact that \citet{muennighoff2023scaling} fit their Chinchilla base law on \citet{hoffmann2022training}'s scraped C4 isoFLOP points rather than on their own scaling runs. Although both studies use C4, differences in tokenization, preprocessing, and training codebases mean that parameters fit on one set of runs need not transfer to another \citep{li2025misfitting}. Indeed, we show that this baseline fits the \citet{muennighoff2023scaling} single-epoch runs poorly, and that the $\widehat{D}, \widehat{N}$ effective-parameter mechanism primarily compensates for this baseline error. When the Chinchilla base is refit on their own data, the two analyses agree.

All results below use the \citet{muennighoff2023scaling} filtered data split (182 runs) with their outlier-removal criteria, matching their evaluation setup.

\para{The Chinchilla baseline transferred from Hoffmann et al.\ fits poorly.} \autoref{tab:chinchilla_baseline_quality} compares the published Chinchilla parameters against parameters refit on the 29 single-epoch runs from \citet{muennighoff2023scaling}. The published parameters explain only 71.1\% of variance on single-epoch data ($R^2 = 0.711$). Since no repetition is involved, these runs should be well-described by a Chinchilla law. Refitting the Chinchilla parameters to the \citet{muennighoff2023scaling} single-epoch data recovers $R^2 = 0.989$, a near-perfect fit. The 3$\times$ reduction in Huber loss confirms that the original parameters are a poor description of these data. The improvement is not driven by relaxing the $\alpha=\beta$ constraint: a tied refit on the same 29 runs also achieves $R^2 = 0.989$.

\para{The effective-parameter mechanism compensates for baseline error.} \autoref{tab:dprime_nprime_comparison} shows the impact on the $\widehat{D}, \widehat{N}$ scaling law. With the published Chinchilla base, adding the $\widehat{D}, \widehat{N}$ mechanism improves $R^2$ by $+0.346$ (from 0.445 to 0.791)---appearing to provide substantial value. With the refit base, the same mechanism improves $R^2$ by only $+0.070$ (from 0.861 to 0.931). The absolute fit quality improves (Huber 0.0158 $\to$ 0.00720), but the marginal contribution of the $\widehat{D}, \widehat{N}$ mechanism is smaller once the baseline is correct.

The mechanistic explanation is visible in the fitted parameters. Under the published base, the effective-parameter rate constant $R_N^* = 5.31$, meaning effective model size saturates quickly, biasing the optimal allocation toward smaller models. Under the refit base, $R_N^* = 3{,}294$, effectively infinite. The optimizer disables the $\widehat{N}$ mechanism entirely. Without $\widehat{N}$ pulling the allocation toward smaller models, the refit $\widehat{D}, \widehat{N}$ law recommends larger models with fewer epochs which aligns with our independent finding.

\begin{figure}[t]
    \centering
    \includegraphics[width=\textwidth]{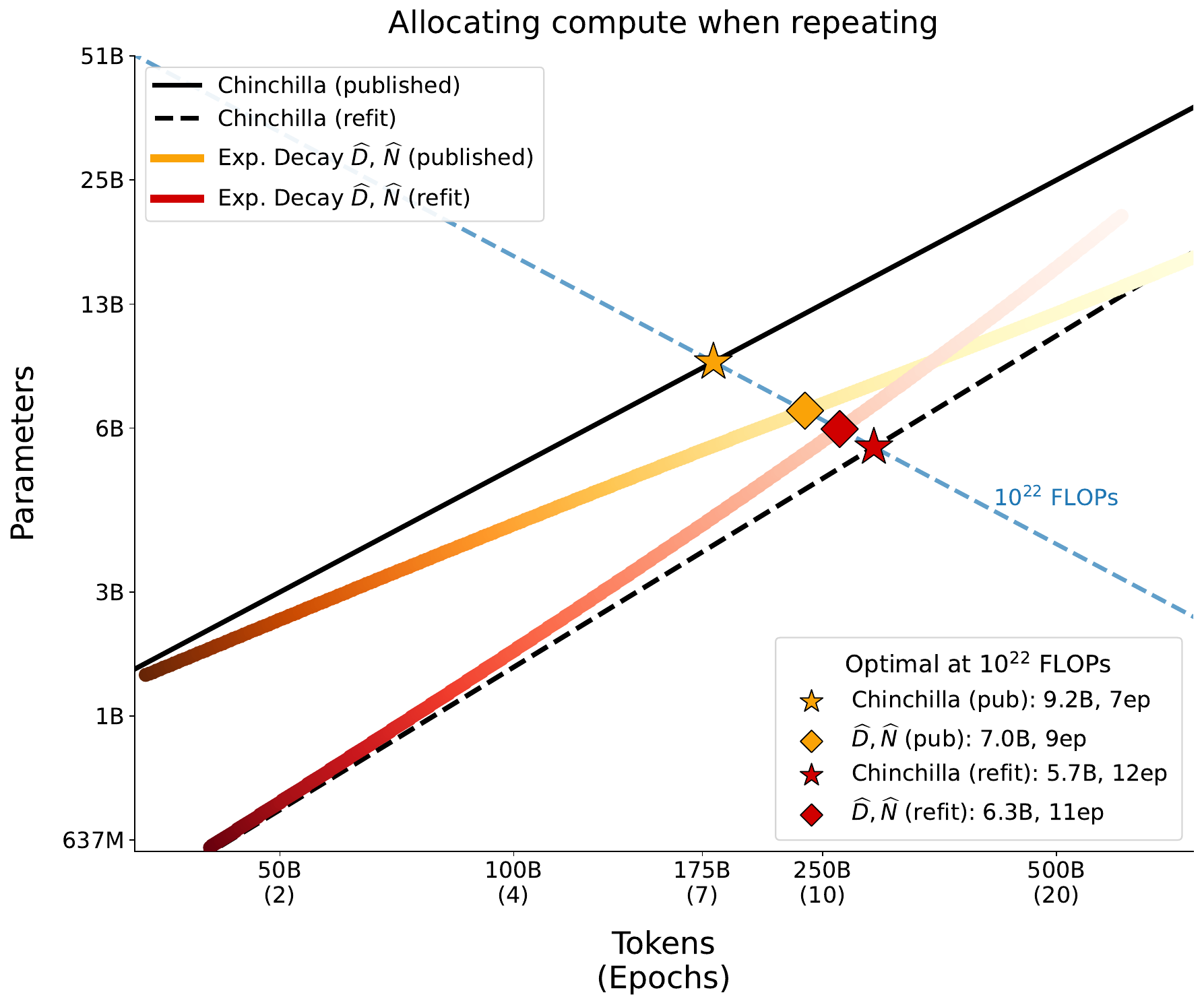}
    \caption{\textbf{Compute-optimal allocation reverses under refit Chinchilla base.} Black lines: Chinchilla-optimal (no repetition). Solid lines: $\widehat{D}, \widehat{N}$-optimal. Yellow: published parameters from \citet{muennighoff2023scaling}. Red: refit parameters. The published $\widehat{D}, \widehat{N}$ law allocates toward smaller models and more data than Chinchilla, while the refit law allocates toward larger models, reversing the original recommendation. $U_D = 25$B tokens.}
    \label{fig:muennighoff_reanalysis_allocation}
\end{figure}

\para{The allocation recommendation reverses.} \autoref{fig:muennighoff_reanalysis_allocation} shows the compute-optimal allocation under the published and refit $\widehat{D}, \widehat{N}$ laws at a unique data budget of 25B tokens. The published law (yellow) allocates toward smaller models and more epochs than Chinchilla, consistent with the original recommendation. The refit law (red) reverses direction, allocating toward larger models---consistent with our finding that compute is better spent on model capacity than on data repetition.

\begin{table}[t]
    \centering
    \caption{\textbf{Chinchilla baseline quality on \citet{muennighoff2023scaling} data.} Published parameters (fit on \citet{hoffmann2022training}'s scraped isoFLOP points) vs.\ parameters refit on the 29 single-epoch runs from \citet{muennighoff2023scaling}.}
    \label{tab:chinchilla_baseline_quality}
    \begin{tabular}{lcc}
        \toprule
        Metric & Published & Refit \\
        \midrule
        $R^2$ (all runs) & 0.445 & \textbf{0.861} \\
        $R^2$ (single-epoch) & 0.711 & \textbf{0.989} \\
        $R^2$ (multi-epoch) & 0.306 & \textbf{0.795} \\
        Huber loss & 0.0331 & \textbf{0.0115} \\
        \bottomrule
    \end{tabular}
\end{table}

\begin{table}[t]
    \centering
    \caption{\textbf{Benefit of $\widehat{D}, \widehat{N}$ under published vs.\ refit Chinchilla base.} $\Delta R^2$ is the improvement over the corresponding Chinchilla-only baseline.}
    \label{tab:dprime_nprime_comparison}
    \begin{tabular}{lccccc}
        \toprule
        Condition & $R^2$ (all) & $\Delta R^2$ & $R^2$ (single) & $R^2$ (multi) & Huber Loss $\downarrow$ \\
        \midrule
        Published Chinchilla & 0.445 & --- & 0.711 & 0.306 & 0.0331 \\
        \quad{}$+\;\widehat{D}, \widehat{N}$ (published) & 0.772 & $+0.327$ & 0.763 & 0.777 & 0.0158 \\
        \quad{}$+\;\widehat{D}, \widehat{N}$ (refit) & 0.791 & $+0.346$ & 0.789 & 0.792 & 0.0158 \\
        \midrule
        Refit Chinchilla  & 0.861 & --- & 0.989 & 0.795 & 0.0115 \\
        \quad{}$+\;\widehat{D}, \widehat{N}$ (refit) & 0.931 & $+0.070$ & 0.989 & 0.902 & 0.00720 \\
        \bottomrule
    \end{tabular}
\end{table}

In summary, the apparent disagreement between our recommendation and that of \citet{muennighoff2023scaling} reduces to the $\widehat{N}$ mechanism absorbing systematic error from a misspecified Chinchilla baseline. When the baseline is corrected, the $\widehat{N}$ mechanism disables itself and the allocation recommendation aligns with ours. We note that this reanalysis was possible only because \citet{muennighoff2023scaling} made their data and experimental details publicly available.

%% file: sections/appendix_training.tex
\section{Training details}
\label{app:training_details}

\subsection{Architecture}

All models follow the Llama~2 architecture~\citep{touvron2023llama} with multi-head attention (MHA) using Rotary Position Embeddings (RoPE), SwiGLU feed-forward networks, and RMSNorm pre-normalization. We use Flash Attention~2 and enable \texttt{torch.compile} for all runs. The vocabulary size is 32,000 tokens (Llama~2 tokenizer).

\subsection{Model configurations}

Table~\ref{tab:model_configs} lists the nine model sizes used in the scaling study. Table~\ref{tab:validation_models} lists eleven additional configurations used only for held-out validation of the fitted scaling law.

\begin{table}[t]
\centering
\caption{Model configurations for the scaling study. Total parameters include embeddings}
\label{tab:model_configs}
\small
\begin{tabular}{lrrrrrrr}
\toprule
Name & $d_\text{model}$ & Layers & Heads & $d_\text{head}$ & $d_\text{ff}$ & Total Params \\
\midrule
15M  & 192   & 6  & 3  & 64  & 768   & 15.87M    \\
25M  & 256   & 8  & 4  & 64  & 1,024 & 24.83M    \\
35M  & 320   & 9  & 5  & 64  & 1,280 & 35.30M    \\
50M  & 384   & 10 & 6  & 64  & 1,536 & 48.25M    \\
125M & 768   & 12 & 12 & 64  & 2,048 & 134.26M   \\
250M & 1,024 & 14 & 16 & 64  & 2,816 & 245.60M   \\
500M & 1,280 & 20 & 10 & 128 & 3,584 & 488.55M   \\
750M & 1,536 & 24 & 12 & 128 & 4,096 & 778.16M   \\
1B   & 1,792 & 26 & 14 & 128 & 4,864 & 1,128.98M \\
\bottomrule
\end{tabular}
\end{table}

\begin{table}[t]
\centering
\caption{Validation-only model configurations, used exclusively for held-out evaluation of the fitted scaling law. The 500M and 1B configurations also used there are listed in \autoref{tab:model_configs}. Names follow the labels in \autoref{tab:held_out_validation} and \autoref{tab:wd_validation}; the Total Params column gives the exact count.}
\label{tab:validation_models}
\small
\begin{tabular}{lrrrrrr}
\toprule
Name & $d_\text{model}$ & Layers & Heads & $d_\text{head}$ & $d_\text{ff}$ & Total Params\\
\midrule
280M & 1,024 & 17 & 16 & 64  & 2,816 & 284.14M \\
350M & 1,024 & 24 & 8  & 128 & 2,816 & 374.07M \\
390M & 1,152 & 20 & 9  & 128 & 3,072 & 392.51M \\
550M & 1,280 & 23 & 10 & 128 & 3,584 & 549.51M \\
700M & 1,408 & 24 & 11 & 128 & 4,096 & 696.02M \\
830M & 1,536 & 26 & 12 & 128 & 4,096 & 834.79M \\
950M & 1,536 & 30 & 12 & 128 & 4,096 & 948.05M \\
2.2B & 2,304 & 32 & 18 & 128 & 6,144 & 2,186.50M \\
2.5B & 2,560 & 30 & 20 & 128 & 6,912 & 2,543.46M \\
3B   & 2,688 & 32 & 21 & 128 & 6,912 & 2,881.22M \\
5B   & 3,328 & 36 & 26 & 128 & 8,960 & 5,029.22M \\
\bottomrule
\end{tabular}
\end{table}

\subsection{Training hyperparameters}

Table~\ref{tab:hyperparams} summarizes the training hyperparameters shared across all runs. The learning rate follows a cosine schedule that decays from the peak value to 10\% of peak over the full training duration, with linear warmup over the first 1\% of steps.

\begin{table}[t]
\centering
\caption{Training hyperparameters.}
\label{tab:hyperparams}
\begin{tabular}{ll}
\toprule
Parameter & Value \\
\midrule
Optimizer & AdamW \\
Learning rate & $2 \times 10^{-4}$ \\
$\beta_1, \beta_2$ & 0.9, 0.95 \\
Weight decay (standard) & 0.1 \\
Weight decay (strong) & 1.0 \\
Gradient clipping & 1.0 (max norm) \\
LR schedule & Cosine with linear warmup \\
Min LR ratio & 0.1 \\
Warmup ratio & 0.01 \\
Sequence length & 512 \\
Batch size & 128 \\
Mixed precision & bf16 \\
\bottomrule
\end{tabular}
\end{table}

\subsection{Dataset}

We train on FineWeb sample-10BT~\citep{penedo2024fineweb}, a 10 billion token sample of the FineWeb corpus. Documents are packed into fixed-length chunks of 512 tokens with end-of-sequence token separators.

\subsection{Evaluation}

We report the final validation loss computed over a held-out set of 1{,}000 documents, each truncated to the 512-token sequence length; no early stopping is applied.

\subsection{Infrastructure}

All experiments are run on NVIDIA B200 GPUs or NVIDIA A6000 GPUs, with one GPU per run.

%% file: sections/appendix_grid.tex
\section{Experimental grid}
\label{app:experimental_grid}

This appendix details the full experimental grid for both the standard ($\lambda = 0.1$) and strong weight decay ($\lambda = 1.0$) studies. All runs use the Llama~2 architecture trained on FineWeb with a causal language modeling (CLM) objective. The grid spans nine model sizes (15M--1B parameters) and eight unique data budgets (50M--6B tokens), with multi-epoch runs at up to 16 epochs ($R_D \in \{0, 1, 3, 7, 11, 15\}$). Coverage differs slightly between sweeps, with a few cells absent from each (Tables~\ref{tab:grid_std_single}--\ref{tab:grid_wd_multi}, \ref{tab:grid_fwedu_single}, and~\ref{tab:grid_fwedu_multi}). In practice, we did not find this to have a meaningful impact on downstream scaling law fits.

\subsection{Standard study ($\lambda = 0.1$)}

\paragraph{Phase~1: single-epoch runs.}
\autoref{tab:grid_std_single} shows the single-epoch grid. Larger models are restricted to smaller budgets due to compute constraints, yielding a banded design.

\begin{table}[t]
    \centering
    \caption{\textbf{Single-epoch grid, standard weight decay ($\lambda = 0.1$).} A \checkmark\ indicates a completed run. 9 model sizes $\times$ 8 data budgets.}
    \label{tab:grid_std_single}
    \footnotesize
    \begin{tabular}{l cccccccc}
        \toprule
        Model & 50M & 100M & 200M & 400M & 800M & 1.5B & 3B & 6B \\
        \midrule
        15M  & \checkmark & \checkmark & \checkmark & \checkmark &  &  &  &  \\
        25M  & \checkmark & \checkmark & \checkmark & \checkmark & \checkmark & \checkmark & \checkmark & \checkmark \\
        35M  & \checkmark & \checkmark & \checkmark & \checkmark &  &  &  &  \\
        50M  & \checkmark & \checkmark & \checkmark & \checkmark & \checkmark & \checkmark & \checkmark & \checkmark \\
        125M & \checkmark & \checkmark & \checkmark & \checkmark & \checkmark & \checkmark & \checkmark & \checkmark \\
        250M & \checkmark & \checkmark & \checkmark & \checkmark & \checkmark & \checkmark & \checkmark & \checkmark \\
        500M & \checkmark & \checkmark & \checkmark & \checkmark & \checkmark & \checkmark &  &  \\
        750M & \checkmark & \checkmark & \checkmark & \checkmark & \checkmark &  &  &  \\
        1B   & \checkmark & \checkmark & \checkmark & \checkmark & \checkmark &  &  &  \\
        \bottomrule
    \end{tabular}
\end{table}

\paragraph{Phase~2: multi-epoch runs.}
\autoref{tab:grid_std_multi} lists the multi-epoch cells. Epoch counts correspond to repetition levels $R_D = 0, 1, 3, 7, 11, 15$.

\begin{table}[t]
    \centering
    \caption{\textbf{Multi-epoch grid, standard weight decay ($\lambda = 0.1$).} Each row is a (model, $U_D$) cell; columns indicate available epoch counts.}
    \label{tab:grid_std_multi}
    \footnotesize
    \begin{tabular}{ll cccccc}
        \toprule
        Model & $U_D$ & 1 & 2 & 4 & 8 & 12 & 16 \\
        \midrule
        25M  & 50M  &            & \checkmark & \checkmark &            &            & \checkmark \\
        25M  & 100M &            &            & \checkmark & \checkmark & \checkmark & \checkmark \\
        25M  & 200M & \checkmark & \checkmark & \checkmark & \checkmark & \checkmark & \checkmark \\
        \addlinespace
        50M  & 50M  & \checkmark & \checkmark & \checkmark & \checkmark & \checkmark & \checkmark \\
        50M  & 100M & \checkmark & \checkmark & \checkmark & \checkmark & \checkmark & \checkmark \\
        50M  & 200M & \checkmark & \checkmark & \checkmark & \checkmark & \checkmark & \checkmark \\
        \addlinespace
        125M & 100M &            & \checkmark & \checkmark & \checkmark & \checkmark & \checkmark \\
        125M & 200M & \checkmark & \checkmark & \checkmark & \checkmark & \checkmark & \checkmark \\
        125M & 400M & \checkmark & \checkmark & \checkmark & \checkmark & \checkmark & \checkmark \\
        125M & 800M &            & \checkmark & \checkmark & \checkmark & \checkmark & \checkmark \\
        \addlinespace
        250M & 100M & \checkmark & \checkmark & \checkmark & \checkmark & \checkmark & \checkmark \\
        250M & 200M & \checkmark & \checkmark & \checkmark & \checkmark & \checkmark & \checkmark \\
        250M & 400M & \checkmark & \checkmark & \checkmark & \checkmark & \checkmark & \checkmark \\
        250M & 800M &            & \checkmark & \checkmark & \checkmark & \checkmark & \checkmark \\
        \addlinespace
        500M & 100M & \checkmark & \checkmark & \checkmark & \checkmark & \checkmark & \checkmark \\
        500M & 200M & \checkmark & \checkmark & \checkmark & \checkmark &            & \checkmark \\
        500M & 400M & \checkmark & \checkmark & \checkmark & \checkmark & \checkmark & \checkmark \\
        \bottomrule
    \end{tabular}
\end{table}

\subsection{Strong weight decay study ($\lambda = 1.0$)}

The strong weight decay runs were collected as one sweep over
$\lambda \in \{0.5, 1.0, 3.0\}$, with every $\lambda$ trained on an identical
grid so that the fitted coefficients are directly comparable. The tables below
therefore describe all three settings; the $\lambda = 0.5$ and $\lambda = 3.0$
arms are analyzed in \autoref{app:wd_sweep}.

\paragraph{Phase~1: single-epoch runs.}
\autoref{tab:grid_wd_single} shows the single-epoch grid, 53 cells per $\lambda$. Larger models (750M, 1B) and larger budgets for mid-sized models provide only single-epoch runs, anchoring the Chinchilla baseline without multi-epoch extensions.

\begin{table}[t]
    \centering
    \caption{\textbf{Single-epoch grid, strong weight decay sweep.} Identical for $\lambda \in \{0.5, 1.0, 3.0\}$. 9 model sizes $\times$ up to 8 data budgets.}
    \label{tab:grid_wd_single}
    \footnotesize
    \begin{tabular}{l cccccccc}
        \toprule
        Model & 50M & 100M & 200M & 400M & 800M & 1.5B & 3B & 6B \\
        \midrule
        15M  & \checkmark & \checkmark & \checkmark & \checkmark &            &            &            & \\
        25M  & \checkmark & \checkmark & \checkmark & \checkmark & \checkmark & \checkmark & \checkmark & \checkmark \\
        35M  & \checkmark & \checkmark & \checkmark & \checkmark &            &            &            & \\
        50M  & \checkmark & \checkmark & \checkmark & \checkmark & \checkmark & \checkmark & \checkmark & \checkmark \\
        125M &            &            & \checkmark & \checkmark & \checkmark & \checkmark & \checkmark & \checkmark \\
        250M &            &            &            & \checkmark & \checkmark & \checkmark & \checkmark & \checkmark \\
        500M & \checkmark & \checkmark & \checkmark & \checkmark & \checkmark & \checkmark & \checkmark & \checkmark \\
        750M & \checkmark & \checkmark & \checkmark & \checkmark & \checkmark &            &            & \\
        1B   & \checkmark & \checkmark & \checkmark & \checkmark & \checkmark &            &            & \\
        \bottomrule
    \end{tabular}
\end{table}

\paragraph{Phase~2: multi-epoch runs.}
\autoref{tab:grid_wd_multi} lists the multi-epoch cells for the strong weight decay study.

\begin{table}[t]
    \centering
    \caption{\textbf{Multi-epoch grid, strong weight decay sweep.} Identical for $\lambda \in \{0.5, 1.0, 3.0\}$.}
    \label{tab:grid_wd_multi}
    \footnotesize
    \begin{tabular}{ll cccccc}
        \toprule
        Model & $U_D$ & 1 & 2 & 4 & 8 & 12 & 16 \\
        \midrule
        25M  & 50M   & \checkmark & \checkmark & \checkmark & \checkmark & \checkmark & \checkmark \\
        25M  & 100M  & \checkmark & \checkmark & \checkmark & \checkmark & \checkmark & \checkmark \\
        25M  & 200M  & \checkmark & \checkmark & \checkmark & \checkmark & \checkmark & \checkmark \\
        \addlinespace
        50M  & 50M   & \checkmark & \checkmark & \checkmark & \checkmark & \checkmark & \checkmark \\
        50M  & 100M  & \checkmark & \checkmark & \checkmark & \checkmark & \checkmark & \checkmark \\
        50M  & 200M  & \checkmark & \checkmark & \checkmark & \checkmark & \checkmark & \checkmark \\
        \addlinespace
        125M & 100M  &            & \checkmark & \checkmark & \checkmark & \checkmark & \checkmark \\
        125M & 200M  & \checkmark & \checkmark & \checkmark & \checkmark & \checkmark & \checkmark \\
        125M & 400M  & \checkmark & \checkmark & \checkmark & \checkmark & \checkmark & \checkmark \\
        125M & 800M  & \checkmark & \checkmark & \checkmark & \checkmark & \checkmark & \checkmark \\
        \addlinespace
        250M & 100M  &            & \checkmark & \checkmark & \checkmark & \checkmark & \checkmark \\
        250M & 200M  &            & \checkmark & \checkmark & \checkmark & \checkmark & \checkmark \\
        250M & 400M  & \checkmark & \checkmark & \checkmark & \checkmark & \checkmark & \checkmark \\
        250M & 800M  & \checkmark & \checkmark & \checkmark & \checkmark & \checkmark & \checkmark \\
        \addlinespace
        500M & 100M  & \checkmark & \checkmark & \checkmark & \checkmark & \checkmark & \checkmark \\
        500M & 200M  & \checkmark & \checkmark & \checkmark & \checkmark & \checkmark & \checkmark \\
        500M & 400M  & \checkmark & \checkmark & \checkmark & \checkmark & \checkmark & \checkmark \\
        \bottomrule
    \end{tabular}
\end{table}

%% file: sections/appendix_fits.tex
\section{Scaling law fit details}
\label{app:fit_details}

This appendix reports the full numerical results for all scaling law fits discussed in the main text, including Phase~1 (Chinchilla baseline) and Phase~2 (repetition penalty) parameter estimates, goodness-of-fit metrics, and external validation on the \citet{muennighoff2023scaling} C4 dataset.

\subsection{Fitting methodology}

All scaling laws are fit by minimizing Huber loss in log-space with $\delta_{\text{Huber}} = 10^{-3}$, using L-BFGS with strong Wolfe line search conditions.
To mitigate sensitivity to initialization, we employ grid search: 324 initializations for the Chinchilla base law (a $4 \times 3 \times 3 \times 3 \times 3$ grid over $E, A, \alpha, B, \beta$) and 7--320 initializations for the repetition penalty forms depending on the number of free parameters.

We report three variants of the coefficient of determination:
\begin{itemize}
    \item $R^2$: standard $1 - \text{SS}_{\text{res}} / \text{SS}_{\text{tot}}$, computed over all data points.
    \item $R^2_{\text{multi}}$: $R^2$ restricted to multi-epoch runs only (epochs $> 1$), isolating the model's ability to capture repetition effects.
    \item $R^2_{\text{single}}$: $R^2$ restricted to single-epoch runs ($R_D = 0$). 
\end{itemize}

Repetition is parameterized as $R_D = \text{epochs} - 1$ with $U_D$ denoting the unique token budget.
Phase~1 parameters ($E, A, \alpha, B, \beta$) are locked during Phase~2 fitting; only the repetition penalty parameters are optimized.
All fits use total parameter counts (embedding + non-embedding).
Uncertainties are reported as the scaled median absolute deviation ($1.4826 \times$ MAD) across 250 bootstrap resamples. Runs are resampled with replacement within each model size, Phase~1 parameters are held fixed, and every resample is refit from the full initialization grid.
Two parameters are reported without uncertainty. $R^*_N$ is not determined by the data: its correction vanishes as $R^*_N \to \infty$, so all large values fit equally well. In the four-parameter form, $P$ and the exponents are determined jointly. $P$ sets the penalty's scale at $N = U_D = 1$, far outside the data, so a small shift in the exponents is offset by a compensating change in $P$, much as the Chinchilla coefficient $A$ co-varies with $\alpha$. The penalty itself is pinned down closely. For example, in the standard study at $N = 250$M, $U_D = 200$M and 8 epochs the additive penalty is $0.135 \pm 0.014$ nats.

\subsection{Phase 1: Chinchilla parameters}

Table~\ref{tab:chinchilla_params} reports the Chinchilla baseline parameters for both weight decay conditions.
The base law takes the form $L(N, D) = E + A / N^\alpha + B / D^\beta$, fit to single-epoch runs only.
Strong weight decay ($\lambda = 1.0$) raises the entropy floor $E$ by ${\sim}0.2$~nats and significantly increases $B$ with a steeper data exponent $\beta$, consistent with stronger regularization reducing effective model capacity at fixed parameter count.

\begin{table}[t]
\centering
\caption{Phase~1 Chinchilla parameters for the standard ($\lambda = 0.1$) and strong weight decay ($\lambda = 1.0$) studies.}
\label{tab:chinchilla_params}
\begin{tabular}{lcc}
\toprule
Parameter & Standard ($\lambda = 0.1$, $n = 60$) & Strong WD ($\lambda = 1.0$, $n = 53$) \\
\midrule
$E$      & 1.8383 & 2.0457 \\
$A$      & 216.58 & 232.72 \\
$\alpha$ & 0.2999 & 0.2973 \\
$B$      & 4{,}964.42 & 25{,}509.95 \\
$\beta$  & 0.4274 & 0.5254 \\
\midrule
$R^2$    & 0.9977 & 0.9949 \\
\bottomrule
\end{tabular}
\end{table}

\subsection{Phase 2: Repetition penalty parameters}

Phase~2 fits the repetition penalty on top of the locked Phase~1 parameters.
We compare five functional forms: two baselines from prior work and three additive penalty forms from our complexity ladder.

\paragraph{Baseline forms.}
The \emph{exponential decay} form from \citet{muennighoff2023scaling} defines effective data $\widehat{D} = U_D \cdot (1 + R^* \cdot (1 - \exp(-R_D / R^*)))$ and evaluates $L(N, \widehat{D}) = E + A/N^\alpha + B/{\widehat{D}}^\beta$.
The \emph{effective parameter} extension adds $\widehat{N} = U_N \cdot (1 + R^*_N \cdot (1 - \exp(-R_N / R^*_N)))$ and evaluates $L(\widehat{N}, \widehat{D})$.

\paragraph{Additive Penalty forms.}
The one-parameter form (\autoref{eq:one_param}) adds a penalty $P \cdot R_D \cdot (N / U_D)$.
The two-parameter form (\autoref{eq:two_param}) generalizes to $P \cdot R_D \cdot (N / U_D)^\kappa$.
The four-parameter form (\autoref{eq:four_param}) further generalizes to $P \cdot R_D^\delta \cdot (N / U_D^\gamma)^\kappa$.

\paragraph{Standard study ($\lambda = 0.1$).}
Table~\ref{tab:phase2_standard} reports fits on our FineWeb data ($n = 153$ total, $n_{\text{multi}} = 81$ multi-epoch runs).
Even the one-parameter additive penalty substantially outperforms both baselines, achieving $R^2_{\text{multi}} = 0.9503$ versus $0.5825$ for the exponential decay form.
The four-parameter additive penalty reaches $R^2 = 0.9970$.

\begin{table}[t]
\centering
\caption{Phase~2 repetition penalty fits for the standard study ($\lambda = 0.1$, $n = 153$, $n_{\text{multi}} = 81$). Best values per metric in bold. Parameters without uncertainty are not separately identified (see text).}
\label{tab:phase2_standard}
\footnotesize
\begin{tabular}{l >{\raggedright\arraybackslash}p{5.4cm} ccc}
\toprule
Form & Parameters (value $\pm$ MAD) & $R^2$ & $R^2_{\text{multi}}$ & Huber \\
\midrule
Exp.\ Decay ($\widehat{D}$) & $R^* = 7.752 \pm 0.954$ & 0.8826 & 0.5825 & 0.003453 \\
Eff.\ Param.\ ($\widehat{D}, \widehat{N}$) & $R^*_D = 7.772 \pm 0.970$, $R^*_N = 4.97 \times 10^{5}$ & 0.8825 & 0.5823 & 0.003451 \\
\midrule
Add.\ Penalty (1p) & $P = 0.02305 \pm 0.00204$ & 0.9847 & 0.9503 & 0.001723 \\
Add.\ Penalty (2p) & $P = 0.02186 \pm 0.00222$, $\kappa = 1.051 \pm 0.188$ & 0.9856 & 0.9533 & 0.001706 \\
Add.\ Penalty (4p) & $P = 3.271 \times 10^{-7}$, $\delta = 1.675 \pm 0.156$, $\kappa = 1.345 \pm 0.0715$, $\gamma = 0.6355 \pm 0.0512$ & \textbf{0.9970} & \textbf{0.9945} & \textbf{0.000976} \\
\bottomrule
\end{tabular}
\end{table}

\paragraph{Strong weight decay study ($\lambda = 1.0$).}
Table~\ref{tab:phase2_strongwd} reports the analogous fits ($n = 138$ total, $n_{\text{multi}} = 85$).
Two observations stand out.
First, the exponential decay baseline performs considerably better under strong WD than under standard WD, reflecting the reduced overfitting when regularization is strong.
Second, the four-parameter additive penalty recovers $\gamma \approx 1.04$ for strong WD versus $0.64$ for standard, indicating that the data-scaling exponent in the penalty term is itself modulated by regularization strength.
The exponential decay form's $R^* = 13.97$ under strong WD versus $7.75$ under standard WD confirms that repetition is more tolerable with strong regularization.

\begin{table}[t]
\centering
\caption{Phase~2 repetition penalty fits for the strong weight decay study ($\lambda = 1.0$, $n = 138$, $n_{\text{multi}} = 85$). Best values per metric in bold. Parameters without uncertainty are not separately identified (see text).}
\label{tab:phase2_strongwd}
\footnotesize
\begin{tabular}{l >{\raggedright\arraybackslash}p{5.4cm} ccc}
\toprule
Form & Parameters (value $\pm$ MAD) & $R^2$ & $R^2_{\text{multi}}$ & Huber \\
\midrule
Exp.\ Decay ($\widehat{D}$) & $R^* = 13.97 \pm 1.88$ & 0.9792 & 0.9476 & 0.001627 \\
Eff.\ Param.\ ($\widehat{D}, \widehat{N}$) & $R^*_D = 14.27 \pm 2.23$, $R^*_N = 4{,}436$ & 0.9793 & 0.9479 & 0.001622 \\
\midrule
Add.\ Penalty (1p) & $P = 0.007071 \pm 0.000527$ & 0.9933 & 0.9888 & 0.001068 \\
Add.\ Penalty (2p) & $P = 0.005843 \pm 0.000533$, $\kappa = 1.332 \pm 0.161$ & 0.9949 & 0.9936 & 0.001015 \\
Add.\ Penalty (4p) & $P = 0.003794$, $\delta = 1.571 \pm 0.0943$, $\kappa = 1.371 \pm 0.0631$, $\gamma = 1.039 \pm 0.0897$ & \textbf{0.9957} & \textbf{0.9959} & \textbf{0.000910} \\
\bottomrule
\end{tabular}
\end{table}

\subsection{External validation on Muennighoff data}

To verify that our penalty forms generalize beyond our experimental setup, we refit all models on the \citet{muennighoff2023scaling}  C4 runs (one run per configuration, without replicates or variants) ($n = 158$, up to 64 epochs).
Phase~1 Chinchilla parameters on this data: $E = 1.9031$, $A = 432.63$, $\alpha = 0.3362$, $B = 5360.24$, $\beta = 0.3868$.

Table~\ref{tab:muennighoff_refit} reports Phase~2 results.
We additionally report $R^2_{\text{single}}$, the coefficient of determination on single-epoch runs, to confirm that the repetition penalty does not degrade the Chinchilla baseline fit.
All forms except $\widehat{D}, \widehat{N}$ achieve identical $R^2_{\text{single}} = 0.9763$ by construction, since the additive penalty terms vanish at $R_D = 0$. The $\widehat{D}, \widehat{N}$ form attains $R^2_{\text{single}} = 0.9832$ because its $\widehat{N}$ saturation modifies predictions for over-parameterized single-epoch runs.
On multi-epoch runs, all three additive penalty forms outperform both baselines, with the four-parameter form achieving the best overall fit ($R^2_{\text{multi}} = 0.9617$).

\begin{table}[t]
\centering
\caption{Repetition penalty fits on the \citet{muennighoff2023scaling} C4 deduplicated data ($n = 158$, up to 64 epochs). Best values per metric in bold.}
\label{tab:muennighoff_refit}
\small
\begin{tabular}{llcccc}
\toprule
Form & Key Params & $R^2$ & $R^2_{\text{multi}}$ & $R^2_{\text{single}}$ & Huber \\
\midrule
Exp.\ Decay ($\widehat{D}$) & $R^* = 23.82$ & 0.8953 & 0.8442 & 0.9763 & 0.008239 \\
Eff.\ Param.\ ($\widehat{D}, \widehat{N}$) & $R^*_D = 38.71$, $R^*_N = 288.1$ & 0.9119 & 0.8670 & 0.9832 & 0.007987 \\
\midrule
Add.\ Penalty (1p) & $P = 0.002857$ & 0.9557 & 0.9426 & 0.9763 & 0.005910 \\
Add.\ Penalty (2p) & $P = 0.006670$, $\kappa = 0.582$ & 0.9633 & 0.9549 & 0.9763 & 0.005528 \\
Add.\ Penalty (4p) & \makecell[l]{$P = 2.48 \times 10^{-6}$, $\delta = 1.040$, \\ $\kappa = 0.803$, $\gamma = 0.526$} & {0.9675} & {0.9617} & {0.9763} & {0.004256} \\
\bottomrule
\end{tabular}
\end{table}

\subsection{FLOPs approximation}

For compute-optimal allocation analysis, we approximate training FLOPs as
\begin{equation}
    C_{\text{train}} \approx 6 \cdot N \cdot D_{\text{total}} = 6 \cdot N \cdot U_D \cdot (1 + R_D).
\end{equation}
Given a fixed compute budget $C_{\text{train}}$ and unique data budget $U_D$, we sweep over candidate epoch counts, compute the implied model size $N = C_{\text{train}} / (6 \cdot U_D \cdot \text{epochs})$, evaluate the predicted loss under each scaling law, and select the allocation that minimizes loss.

%% file: sections/appendix_fwedu.tex
\section{Generalization to FineWeb-Edu}
\label{app:fineweb_edu}

The fits in the main text are all on FineWeb. To validate the robustness of our finding across datasets,
we re-ran the entire sweep on FineWeb-Edu \citep{penedo2024fineweb},
a higher-quality educational subset of FineWeb, and refit every form from
scratch.

The replication uses the standard study's hyperparameters ($\lambda = 0.1$) on the weight decay sweep's grid (\autoref{tab:grid_fwedu_single}, \autoref{tab:grid_fwedu_multi}). Only the pretraining corpus changes.

\subsection{Phase 1: Chinchilla parameters}

\autoref{tab:fwedu_chinchilla} reports the single-epoch fit.
The filtering lowers the entropy floor $E$, so
FineWeb-Edu models reach lower validation loss at every scale, including in the
single-epoch Chinchilla regime.

\begin{table}[t]
\centering
\caption{Phase~1 Chinchilla parameters on FineWeb-Edu, with the FineWeb CLM fit for reference. The heavier filtering lowers the entropy floor $E$.}
\label{tab:fwedu_chinchilla}
\begin{tabular}{lcc}
\toprule
Parameter & FineWeb ($n = 60$) & FineWeb-Edu ($n = 53$) \\
\midrule
$E$      & 1.8383 & 1.6857 \\
$A$      & 216.58 & 601.04 \\
$\alpha$ & 0.2999 & 0.3622 \\
$B$      & 4{,}964.42 & 8{,}775.68 \\
$\beta$  & 0.4274 & 0.4593 \\
\midrule
$R^2$    & 0.9977 & 0.9976 \\
\bottomrule
\end{tabular}
\end{table}

\subsection{Phase 2: repetition penalty}

\autoref{tab:fwedu_phase2} reports the same five forms as
\autoref{tab:phase2_standard}, and the ordering reproduces exactly.
Both effective-data baselines fail to track multi-epoch loss
($R^2_{\text{multi}} = 0.6274$ and $0.6277$), while every additive form dominates:
even the one-parameter penalty reaches $0.9500$, and the four-parameter form
reaches $0.9922$ while cutting Huber loss by roughly $70\%$ relative to both
baselines.

\begin{table}[t]
\centering
\caption{Phase~2 repetition penalty fits on FineWeb-Edu ($n = 135$, $n_{\text{multi}} = 82$). Best values per metric in bold. Parameters without uncertainty are not separately identified (\autoref{app:fit_details}).}
\label{tab:fwedu_phase2}
\footnotesize
\begin{tabular}{l >{\raggedright\arraybackslash}p{5.4cm} ccc}
\toprule
Form & Parameters (value $\pm$ MAD) & $R^2$ & $R^2_{\text{multi}}$ & Huber \\
\midrule
Exp.\ Decay ($\widehat{D}$) & $R^* = 7.455 \pm 2.21$ & 0.8756 & 0.6274 & 0.004065 \\
Eff.\ Param.\ ($\widehat{D}, \widehat{N}$) & $R^*_D = 7.419 \pm 2.19$, $R^*_N = 7.341 \times 10^{6}$ & 0.8757 & 0.6277 & 0.004038 \\
\midrule
Add.\ Penalty (1p) & $P = 0.02182 \pm 0.000372$ & 0.9821 & 0.9500 & 0.002065 \\
Add.\ Penalty (2p) & $P = 0.02111 \pm 0.00174$, $\kappa = 1.028 \pm 0.120$ & 0.9825 & 0.9512 & 0.002059 \\
Add.\ Penalty (4p) & $P = 9.664 \times 10^{-8}$, $\delta = 1.719 \pm 0.107$, $\kappa = 1.381 \pm 0.0418$, $\gamma = 0.6055 \pm 0.0388$ & \textbf{0.9960} & \textbf{0.9922} & \textbf{0.001217} \\
\bottomrule
\end{tabular}
\end{table}

\subsection{Repetition tolerance is similar}

Because the one-parameter form isolates overfitting in a single coefficient, we
can read repetition tolerance off directly. FineWeb-Edu gives $P = 0.02182$
against FineWeb's $0.02305$. We observe that changing the corpus therefore moves $P$ by $5\%$, where changing the weight decay moves it by a factor of $29$ (\autoref{tab:wd_sweep_penalty}). The four-parameter exponents are likewise close. Repetition damage is superlinear in $R_D$ by
almost exactly the same margin on both corpora, and overfitting sets in at
comparable repetition counts.

\subsection{Experimental grid}

\autoref{tab:grid_fwedu_single} and \autoref{tab:grid_fwedu_multi} detail the single- and multi-epoch grids.

\input{sections/appendix_fwedu_grid}

%% file: sections/appendix_fwedu_grid.tex
\begin{table}[t]
    \centering
    \caption{\textbf{Single-epoch grid, FineWeb-Edu.}}
    \label{tab:grid_fwedu_single}
    \footnotesize
    \begin{tabular}{l cccccccc}
        \toprule
        Model & 50M & 100M & 200M & 400M & 800M & 1.5B & 3B & 6B \\
        \midrule
        15M  & \checkmark & \checkmark & \checkmark & \checkmark &            &            &            & \\
        25M  & \checkmark & \checkmark & \checkmark & \checkmark & \checkmark & \checkmark & \checkmark & \checkmark \\
        35M  & \checkmark & \checkmark & \checkmark & \checkmark &            &            &            & \\
        50M  & \checkmark & \checkmark & \checkmark & \checkmark & \checkmark & \checkmark & \checkmark & \checkmark \\
        125M &            &            & \checkmark & \checkmark & \checkmark & \checkmark & \checkmark & \checkmark \\
        250M &            &            &            & \checkmark & \checkmark & \checkmark & \checkmark & \checkmark \\
        500M & \checkmark & \checkmark & \checkmark & \checkmark & \checkmark & \checkmark & \checkmark & \checkmark \\
        750M & \checkmark & \checkmark & \checkmark & \checkmark & \checkmark &            &            & \\
        1B   & \checkmark & \checkmark & \checkmark & \checkmark & \checkmark &            &            & \\
        \bottomrule
    \end{tabular}
\end{table}

\begin{table}[t]
    \centering
    \caption{\textbf{Multi-epoch grid, FineWeb-Edu.} }
    \label{tab:grid_fwedu_multi}
    \footnotesize
    \begin{tabular}{ll cccccc}
        \toprule
        Model & $U_D$ & 1 & 2 & 4 & 8 & 12 & 16 \\
        \midrule
        25M  & 50M   & \checkmark & \checkmark & \checkmark & \checkmark & \checkmark & \checkmark \\
        25M  & 100M  & \checkmark & \checkmark & \checkmark & \checkmark & \checkmark & \checkmark \\
        25M  & 200M  & \checkmark & \checkmark & \checkmark & \checkmark & \checkmark & \checkmark \\
        \addlinespace
        50M  & 50M   & \checkmark & \checkmark & \checkmark & \checkmark & \checkmark & \\
        50M  & 100M  & \checkmark &            & \checkmark & \checkmark &            & \checkmark \\
        50M  & 200M  & \checkmark & \checkmark & \checkmark & \checkmark & \checkmark & \checkmark \\
        \addlinespace
        125M & 100M  &            & \checkmark & \checkmark & \checkmark & \checkmark & \checkmark \\
        125M & 200M  & \checkmark & \checkmark & \checkmark & \checkmark & \checkmark & \checkmark \\
        125M & 400M  & \checkmark & \checkmark & \checkmark & \checkmark & \checkmark & \checkmark \\
        125M & 800M  & \checkmark & \checkmark & \checkmark & \checkmark & \checkmark & \checkmark \\
        \addlinespace
        250M & 100M  &            & \checkmark & \checkmark & \checkmark & \checkmark & \checkmark \\
        250M & 200M  &            & \checkmark & \checkmark & \checkmark & \checkmark & \checkmark \\
        250M & 400M  & \checkmark & \checkmark & \checkmark & \checkmark & \checkmark & \checkmark \\
        250M & 800M  & \checkmark & \checkmark & \checkmark & \checkmark & \checkmark & \checkmark \\
        \addlinespace
        500M & 100M  & \checkmark & \checkmark & \checkmark & \checkmark & \checkmark & \checkmark \\
        500M & 200M  & \checkmark & \checkmark & \checkmark & \checkmark & \checkmark & \checkmark \\
        500M & 400M  & \checkmark & \checkmark & \checkmark & \checkmark & \checkmark & \checkmark \\
        \bottomrule
    \end{tabular}
\end{table}

%% file: sections/appendix_wd_sweep.tex
\section{The weight decay sweep}
\label{app:wd_sweep}

The case study in \autoref{sec:results} contrasts two weight decay settings,
$\lambda = 0.1$ and $\lambda = 1.0$. This appendix extends the
sweep to $\lambda \in \{0.1, 0.5, 1.0, 3.0\}$.

We train $\lambda \in {0.5, 1.0, 3.0}$ on an identical grid (\autoref{tab:grid_wd_single}, \autoref{tab:grid_wd_multi}). The $\lambda = 0.1$ arm draws on the standard study of \autoref{app:experimental_grid}, whose grid differs slightly. To ensure that all four settings are fit identically, we refit its Phase~1 parameters here. They agree closely with those of \autoref{tab:chinchilla_params}.

\subsection{Phase 1: the single-epoch premium}

\autoref{tab:wd_sweep_chinchilla} fits the Chinchilla law separately per
$\lambda$. The entropy floor $E$ climbs monotonically, from 1.8384 nats at
$\lambda = 0.1$ to 2.7879 at $\lambda = 3.0$, and $B$ grows by over two
orders of magnitude across the range. Stronger regularization is strictly worse
when data is abundant: it pays a loss premium at every compute budget in the
single-epoch regime. This premium is what the lower overfitting coefficient has
to overcome.

\begin{table}[t]
\centering
\caption{Phase~1 Chinchilla parameters across the weight decay sweep. Stronger regularization raises the entropy floor $E$, the single-epoch loss premium that offsets its lower overfitting coefficient.}
\label{tab:wd_sweep_chinchilla}
\begin{tabular}{lcccc}
\toprule
Parameter & $\lambda = 0.1$ ($n = 60$) & $\lambda = 0.5$ ($n = 53$) & $\lambda = 1.0$ ($n = 53$) & $\lambda = 3.0$ ($n = 53$) \\
\midrule
$E$ & 1.8384 & 1.9111 & 2.0457 & 2.7879 \\
$A$ & 216.79 & 258.49 & 232.72 & 962.98 \\
$\alpha$ & 0.3000 & 0.3078 & 0.2973 & 0.3715 \\
$B$ & 4{,}964.47 & 9{,}035.06 & 25{,}509.95 & 1{,}405{,}701.25 \\
$\beta$ & 0.4274 & 0.4628 & 0.5254 & 0.7702 \\
\midrule
$R^2$ & 0.9977 & 0.9966 & 0.9949 & 0.9634 \\
\bottomrule
\end{tabular}
\end{table}

\subsection{Phase 2: the overfitting coefficient falls with $\lambda$}

Because the one-parameter form (\autoref{eq:one_param}) isolates overfitting in a
single coefficient, fitting it per $\lambda$ turns regularization strength into a
single readable number. \autoref{tab:wd_sweep_penalty} reports the result: $P$
falls smoothly and monotonically, by a factor of $29$ from $\lambda = 0.1$ to
$\lambda = 3.0$. The fit quality remains high across all settings.

\begin{table}[t]
\centering
\caption{\textbf{The overfitting coefficient falls monotonically with weight decay.} One-parameter additive penalty (\autoref{eq:one_param}) fit independently per $\lambda$. $P$ is reported $\pm$ its scaled MAD over 250 bootstrap resamples (\autoref{app:fit_details}).}
\label{tab:wd_sweep_penalty}
\small
\begin{tabular}{lcccc}
\toprule
$\lambda$ & $P$ & $P / P(\lambda = 0.1)$ & $R^2_{\text{multi}}$ & Huber \\
\midrule
$0.1$ & $0.02305 \pm 0.0020$ & $1.000\times$ & 0.9503 & 0.001723 \\
$0.5$ & $0.01286 \pm 0.00044$ & $0.558\times$ & 0.9737 & 0.001293 \\
$1.0$ & $0.007071 \pm 0.00053$ & $0.307\times$ & 0.9888 & 0.001068 \\
$3.0$ & $0.000806 \pm 0.00020$ & $0.035\times$ & 0.9716 & 0.001930 \\
\bottomrule
\end{tabular}
\end{table}

\subsection{Compute-optimal weight decay}

The two effects pull in opposite directions: stronger $\lambda$ costs loss in the
single-epoch regime but buys tolerance to repetition. We find that neither dominates, and the
compute-optimal $\lambda$ is regime-dependent. Our law predicts which regime
you are in.

\begin{figure}[t]
    \centering
    \includegraphics[width=\textwidth]{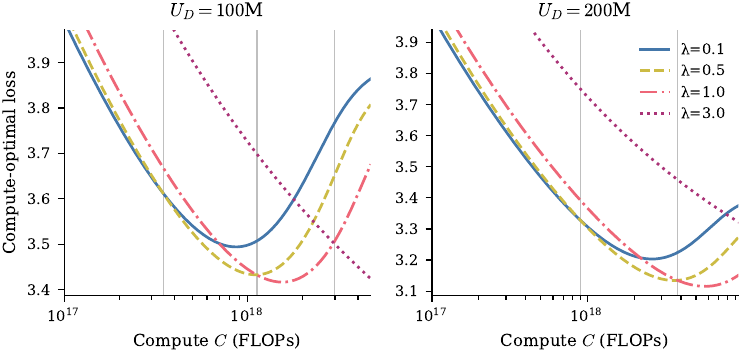}
    \caption{\textbf{The compute-optimal weight decay rises with compute and with tighter data constraints.} Compute-optimal loss under the four-parameter law for each $\lambda$, at two unique data budgets. Vertical lines mark the crossover budgets of \autoref{tab:wd_sweep_crossover}.}
    \label{fig:wd_sweep_frontiers}
\end{figure}

\autoref{fig:wd_sweep_frontiers} plots the compute-optimal loss for each setting
at two unique data budgets, minimizing the four-parameter law
(\autoref{eq:four_param}) over epoch count at every compute budget. The lower envelope sweeps through the
settings in order: $\lambda = 0.1$ is optimal at low compute, and successively
stronger regularization takes over as the compute budget grows and the data
constraint becomes tighter.

\autoref{tab:wd_sweep_crossover} reads the crossover budgets off those curves.
Two patterns are visible. The optimum rises with the compute budget, and it is
reached sooner when data is scarcer: at $U_D = 100$M, $\lambda = 1.0$ becomes
optimal at $1.1 \times 10^{18}$ FLOPs, while at $U_D = 400$M the same transition
needs $1.4 \times 10^{19}$, an order of magnitude more compute. At
$U_D = 400$M the schedule saturates at $\lambda = 1.0$: $\lambda = 3.0$ never
becomes optimal within our compute range. The
repetition robustness fails to repay the large single-epoch premium.

\begin{table}[t]
\centering
\caption{\textbf{Crossover budgets $C^{\times}$ for the compute-optimal weight decay.} Training compute above which each $\lambda$ overtakes the previous one, under the four-parameter law. The optimum rises with compute and is reached sooner under tighter data constraints; at $U_D = 400$M it saturates at $\lambda = 1.0$ within our compute range.}
\label{tab:wd_sweep_crossover}
\small
\begin{tabular}{lccc}
\toprule
Transition & $U_D = 100$M & $U_D = 200$M & $U_D = 400$M \\
\midrule
$\lambda = 0.5$ (from $0.1$) & $3.5 \times 10^{17}$ & $9.0 \times 10^{17}$ & $2.7 \times 10^{18}$ \\
$\lambda = 1.0$ (from $0.5$) & $1.1 \times 10^{18}$ & $3.8 \times 10^{18}$ & $1.4 \times 10^{19}$ \\
$\lambda = 3.0$ (from $1.0$) & $3.0 \times 10^{18}$ & $1.6 \times 10^{19}$ & --- \\
\bottomrule
\end{tabular}
\end{table}

We find that there is thus no single best weight decay. However, our law provides
prescriptive guidance: given $U_D$ and a compute budget, it names the $\lambda$ that
minimizes loss, and the budget at which that answer changes.

%% file: sections/appendix_downstream.tex
\section{Downstream evaluation details}
\label{app:downstream}

This appendix provides evaluation methodology and task-level results supporting the aggregate OLMES Avg.\ BPB scores reported in \autoref{tab:held_out_validation} and \autoref{tab:wd_validation}.

\subsection{Evaluation setup}

All downstream evaluations use the Open Language Model Evaluation Standard (OLMES). We evaluate every task with its standard OLMES configuration (format, few-shot examples, and evaluation split as registered), changing only the maximum sequence length to 512 tokens. We report bits-per-byte (BPB) because it remains informative even when models perform near chance on multiple-choice tasks, as is common at the model scales in our study \citep{heineman2025signal}.

\subsection{Task inventory}

\autoref{tab:task_inventory} lists the 19 tasks used to compute the OLMES Avg.\ BPB.

\begin{table}[t]
    \centering
    \caption{\textbf{Downstream evaluation tasks.}}
    \label{tab:task_inventory}
    \begin{tabular}{ll}
        \toprule
        Benchmark & Source \\
        \midrule
        arc\_easy & \citet{allenai:arc}\\
        arc\_challenge & \citet{allenai:arc}\\
        csqa & \citet{talmor2019commonsenseqa}\\
        hellaswag & \citet{zellers2019hellaswag}\\
        winogrande & \citet{sakaguchi2021winogrande} \\
        socialiqa & \citet{sap2019social}\\
        piqa & \citet{bisk2020piqa}\\
        sciq & \citet{welbl2017crowdsourcing} \\
        qasper\_yesno & \citet{dasigi2021dataset} \\
        lab\_bench\_protocolqa & \citet{laurent2024lab} \\
        medmcqa & \citet{pal2022medmcqa} \\
        medqa\_en & \citet{jin2021disease}\\
        sciriff\_yesno & \citet{wadden2025sciriff} \\
        coqa & \citet{reddy2019coqa} \\
        drop & \citet{dua2019drop} \\
        jeopardy & \citet{kaggle200000jeopardy} \\
        naturalqs\_open & \citet{kwiatkowski-etal-2019-natural} \\
        squad & \citet{rajpurkar2016squad} \\
        lambada & \citet{paperno2016lambada} \\
        \bottomrule
    \end{tabular}
\end{table}

\subsection{Task-level BPB breakdown at $U_D = 250$M}

\autoref{tab:task_bpb_250m} reports per-task BPB for the three matched-compute configurations from the $U_D = 250$M, $C = 5 \times 10^{18}$ comparison in \autoref{tab:held_out_validation}. 

\begin{table}[t]
    \centering
    \caption{\textbf{Per-task BPB at $U_D = 250$M, $C = 5 \times 10^{18}$.} Each column trains the configuration recommended by the corresponding scaling law. Bold indicates the lowest (best) BPB per task.}
    \label{tab:task_bpb_250m}
    \begin{tabular}{lccc}
        \toprule
        Task & Add.\ Penalty (700M) & Chinchilla (280M) & Eff.\ Param.\ (500M) \\
        \midrule
        arc\_easy & {1.3871} & 1.4751 & 1.4091 \\
        arc\_challenge & {1.4711} & 1.5447 & 1.5291 \\
        csqa & {1.7181} & 1.8002 & 1.7786 \\
        hellaswag & {1.0503} & 1.0817 & 1.0591 \\
        winogrande & {1.3739} & 1.4197 & 1.3930 \\
        socialiqa & 1.5555 & 1.6587 & {1.5332} \\
        piqa & {1.3149} & 1.3702 & 1.3576 \\
        sciq & {1.4761} & 1.6006 & 1.5211 \\
        qasper\_yesno & 0.6040 & {0.5763} & 0.7771 \\
        lab\_bench\_protocolqa & {1.7687} & 1.8129 & 1.8001 \\
        medmcqa & {2.1017} & 2.1704 & 2.1507 \\
        medqa\_en & 1.6683 & 1.7082 & {1.6643} \\
        sciriff\_yesno & {0.6709} & 0.8108 & 1.0329 \\
        \midrule
        coqa & {1.2571} & 1.4285 & 1.3463 \\
        drop & 2.3949 & 2.3721 & {2.2548} \\
        jeopardy & {1.7356} & 1.9124 & 1.7891 \\
        naturalqs\_open & {1.7468} & 1.8578 & 1.7868 \\
        squad & {1.2014} & 1.2462 & 1.2563 \\
        \midrule
        lambada & {0.9948} & 1.0356 & 1.0209 \\
        \midrule
        \textbf{Average} & \textbf{1.4469} & 1.5201 & 1.4979 \\
        \bottomrule
    \end{tabular}
\end{table}

\subsection{Per-task BPB at $U_D = 500$M}

\autoref{tab:task_bpb_500m_1e19} and \autoref{tab:task_bpb_500m_2e19} report per-task BPB for the $U_D = 500$M configurations from \autoref{tab:held_out_validation}.

\begin{table}[t]
    \centering
    \caption{\textbf{Per-task BPB at $U_D = 500$M, $C = 1 \times 10^{19}$.} Bold indicates the lowest BPB per task.}
    \label{tab:task_bpb_500m_1e19}
    \small
    \begin{tabular}{lccc}
        \toprule
        Task & Add.\ Penalty (700M) & Chinchilla (390M) & Eff.\ Param.\ (550M) \\
        \midrule
        arc\_easy & {1.2000} & 1.2203 & 1.2640 \\
        arc\_challenge & {1.3275} & 1.3558 & 1.3912 \\
        csqa & {1.4790} & 1.5052 & 1.6061 \\
        hellaswag & 0.9654 & 0.9663 & {0.9626} \\
        winogrande & 1.3264 & 1.3278 & {1.3024} \\
        socialiqa & 1.4118 & {1.3796} & 1.3958 \\
        piqa & 1.2473 & 1.2396 & {1.2348} \\
        sciq & 1.3652 & {1.3373} & 1.3465 \\
        qasper\_yesno & {0.5382} & 0.7592 & 0.6311 \\
        lab\_bench\_protocolqa & 1.6119 & {1.6013} & 1.6494 \\
        medmcqa & 1.9571 & {1.9351} & 1.9450 \\
        medqa\_en & {1.4252} & 1.4696 & 1.4376 \\
        sciriff\_yesno & {0.7209} & 1.0259 & 0.9950 \\
        \midrule
        coqa & {1.1422} & 1.1651 & 1.1568 \\
        drop & {1.9544} & 2.1023 & 2.3233 \\
        jeopardy & 1.5639 & 1.6102 & {1.5423} \\
        naturalqs\_open & {1.5837} & 1.5853 & 1.6106 \\
        squad & {1.0257} & 1.1046 & 1.1294 \\
        \midrule
        lambada & 0.8649 & 0.8646 & {0.8649} \\
        \midrule
        \textbf{Average} & \textbf{1.3006} & 1.3450 & 1.3573 \\
        \bottomrule
    \end{tabular}
\end{table}

\begin{table}[t]
    \centering
    \caption{\textbf{Per-task BPB at $U_D = 500$M, $C = 2 \times 10^{19}$.} Bold indicates the lowest BPB per task.}
    \label{tab:task_bpb_500m_2e19}
    \small
    \begin{tabular}{lccc}
        \toprule
        Task & Add.\ Penalty (2.2B) & Chinchilla (700M) & Eff.\ Param.\ (950M) \\
        \midrule
        arc\_easy & {1.2377} & 1.2786 & 1.3195 \\
        arc\_challenge & {1.3376} & 1.3855 & 1.4071 \\
        csqa & 1.5240 & {1.5207} & 1.6015 \\
        hellaswag & {0.9456} & 1.0056 & 0.9902 \\
        winogrande & {1.3022} & 1.3785 & 1.3588 \\
        socialiqa & 1.4403 & {1.4227} & 1.4787 \\
        piqa & {1.2148} & 1.2938 & 1.2626 \\
        sciq & 1.4716 & {1.3464} & 1.3847 \\
        qasper\_yesno & 0.7131 & {0.6887} & 0.8466 \\
        lab\_bench\_protocolqa & {1.6097} & 1.7216 & 1.6551 \\
        medmcqa & {1.9019} & 2.0671 & 1.9848 \\
        medqa\_en & {1.4514} & 1.5098 & 1.4693 \\
        sciriff\_yesno & {0.9907} & 0.9934 & 1.1969 \\
        \midrule
        coqa & {1.1280} & 1.2350 & 1.1556 \\
        drop & {2.1229} & 2.2267 & 2.3917 \\
        jeopardy & 1.5459 & 1.5913 & {1.5420} \\
        naturalqs\_open & {1.5758} & 1.5812 & 1.5847 \\
        squad & {1.0650} & 1.1494 & 1.1547 \\
        \midrule
        lambada & 0.8431 & 0.8671 & {0.8324} \\
        \midrule
        \textbf{Average} & \textbf{1.3380} & 1.3823 & 1.4009 \\
        \bottomrule
    \end{tabular}
\end{table}

\subsection{Weight decay validation: per-task BPB}

Tables~\ref{tab:task_bpb_wd_250m}--\ref{tab:task_bpb_wd_500m} report per-task BPB for all weight decay validation configurations from \autoref{tab:wd_validation}. The two $\lambda = 0.1$, 700M runs ($U_D = 250$M and 500M) are independent reruns of the corresponding configurations in \autoref{tab:held_out_validation}, so their results differ slightly from those reported there.

\begin{table}[t]
    \centering
    \caption{\textbf{Per-task BPB for weight decay validation at $U_D = 250$M.} Bold indicates the lowest BPB per task within each compute budget.}
    \label{tab:task_bpb_wd_250m}
    \footnotesize
    \begin{tabular}{l cc cc cc}
        \toprule
        & \multicolumn{2}{c}{$C = 3 \times 10^{18}$} & \multicolumn{2}{c}{$C = 5 \times 10^{18}$} & \multicolumn{2}{c}{$C = 1 \times 10^{19}$} \\
        \cmidrule(lr){2-3} \cmidrule(lr){4-5} \cmidrule(lr){6-7}
        Task & \makecell{$\lambda\!=\!0.1$\\350M} & \makecell{$\lambda\!=\!1.0$\\350M} & \makecell{$\lambda\!=\!0.1$\\700M} & \makecell{$\lambda\!=\!1.0$\\550M} & \makecell{$\lambda\!=\!0.1$\\3B} & \makecell{$\lambda\!=\!1.0$\\1B} \\
        \midrule
        arc\_easy & {1.3641} & 1.4125 & 1.4115 & {1.3543} & 1.4498 & {1.2658} \\
        arc\_challenge & {1.4878} & 1.5040 & 1.4912 & {1.4668} & 1.5292 & {1.3788} \\
        csqa & {1.7131} & 1.7456 & 1.6361 & {1.6175} & 1.8680 & {1.5369} \\
        hellaswag & 1.0549 & {1.0361} & 1.0425 & {1.0200} & 1.0445 & {0.9958} \\
        winogrande & 1.3685 & {1.3394} & 1.3738 & {1.3309} & 1.3642 & {1.3302} \\
        socialiqa & {1.4802} & 1.5296 & 1.5134 & {1.4443} & 1.5999 & {1.4099} \\
        piqa & 1.3326 & {1.3186} & 1.3180 & {1.2817} & 1.3226 & {1.2488} \\
        sciq & {1.5020} & 1.5469 & 1.4817 & {1.4547} & 1.6897 & {1.3767} \\
        qasper\_yesno & 0.7511 & {0.5296} & 0.7219 & {0.5649} & 0.6659 & {0.5165} \\
        lab\_bench & 1.7999 & {1.7374} & 1.7661 & {1.6924} & 1.7698 & {1.6877} \\
        medmcqa & 2.1018 & {2.0807} & 2.1374 & {2.0532} & 2.0995 & {1.9886} \\
        medqa\_en & 1.6807 & {1.6796} & 1.6576 & {1.6358} & 1.6918 & {1.5541} \\
        sciriff\_yesno & 0.9035 & {0.8128} & 0.8904 & {0.9680} & 1.0393 & {0.6944} \\
        \midrule
        coqa & 1.2838 & {1.2423} & 1.3242 & {1.2988} & 1.3089 & {1.1676} \\
        drop & {2.0825} & 2.2720 & 2.3355 & {1.9916} & 2.2124 & {2.1823} \\
        jeopardy & {1.7647} & 1.8612 & 1.7728 & {1.7503} & 1.9061 & {1.6939} \\
        naturalqs\_open & {1.8332} & 1.8559 & 1.7940 & {1.7416} & 1.9459 & {1.6699} \\
        squad & {1.2262} & 1.2456 & 1.2437 & {1.1860} & 1.2510 & {1.1139} \\
        \midrule
        lambada & {1.0503} & 1.1133 & 1.0165 & {1.0270} & 1.0643 & {0.9774} \\
        \midrule
        \textbf{Average} & \textbf{1.4622} & 1.4665 & 1.4699 & \textbf{1.4147} & 1.5170 & \textbf{1.3573} \\
        \bottomrule
    \end{tabular}
\end{table}

\begin{table}[t]
    \centering
    \caption{\textbf{Per-task BPB for weight decay validation at $U_D = 500$M.} Bold indicates the lowest BPB per task within each compute budget.}
    \label{tab:task_bpb_wd_500m}
    \small
    \begin{tabular}{l cc cc}
        \toprule
        & \multicolumn{2}{c}{$C = 1 \times 10^{19}$} & \multicolumn{2}{c}{$C = 3 \times 10^{19}$} \\
        \cmidrule(lr){2-3} \cmidrule(lr){4-5}
        Task & \makecell{$\lambda\!=\!0.1$\\700M} & \makecell{$\lambda\!=\!1.0$\\830M} & \makecell{$\lambda\!=\!0.1$\\5B} & \makecell{$\lambda\!=\!1.0$\\2.5B} \\
        \midrule
        arc\_easy & 1.2825 & {1.2172} & 1.2355 & {1.1403} \\
        arc\_challenge & 1.3782 & {1.3631} & 1.3810 & {1.2863} \\
        csqa & 1.6236 & {1.5031} & 1.5299 & {1.4612} \\
        hellaswag & {0.9634} & 0.9642 & 0.9602 & {0.9241} \\
        winogrande & 1.3076 & {1.3033} & 1.3043 & {1.2700} \\
        socialiqa & 1.4634 & {1.3925} & 1.3804 & {1.3567} \\
        piqa & {1.2343} & 1.2287 & 1.2173 & {1.1868} \\
        sciq & 1.4068 & {1.3495} & 1.3899 & {1.2698} \\
        qasper\_yesno & 0.5958 & {0.5650} & 0.7242 & 0.7816 \\
        lab\_bench\_protocolqa & {1.6147} & 1.6238 & 1.6170 & {1.5449} \\
        medmcqa & 1.9630 & {1.9180} & 1.9346 & {1.8650} \\
        medqa\_en & {1.4841} & 1.4975 & 1.4518 & {1.3914} \\
        sciriff\_yesno & 1.0341 & {0.8033} & 1.0518 & {1.0509} \\
        \midrule
        coqa & {1.1169} & 1.1846 & 1.1699 & {1.0600} \\
        drop & 2.0835 & {2.0114} & {1.9909} & 2.1863 \\
        jeopardy & 1.6474 & {1.6069} & 1.6346 & {1.5870} \\
        naturalqs\_open & 1.6121 & {1.6059} & 1.6174 & {1.5010} \\
        squad & 1.1165 & {1.0818} & 1.1257 & {1.1133} \\
        \midrule
        lambada & {0.8591} & 0.9333 & 0.8821 & {0.8155} \\
        \midrule
        \textbf{Average} & 1.3572 & \textbf{1.3238} & 1.3473 & \textbf{1.3048} \\
        \bottomrule
    \end{tabular}
\end{table}

%% file: sections/appendix_llm.tex
\section{LLM usage}
\label{app:llm-usage}

Large language models were used as general-purpose writing and coding assistants during the development of this work.